\documentclass[11pt,a4paper,table]{article}
\usepackage[hyperref]{emnlp2020}
\usepackage{times}
\usepackage{latexsym}
\usepackage{booktabs}
\usepackage{graphicx}
\usepackage{amsmath}
\usepackage{multirow}
\usepackage{pgfplots}
\usepackage{subcaption}
\usepackage{wrapfig}
\usepackage{hhline}
\usepackage{tabularx}
\usepackage{xcolor}
\usepackage{makecell}
\include{Definitions}

\newcommand{\cmmnt}[1]{\ignorespaces}

\usepackage{microtype}

\aclfinalcopy 

\title{Improving Text Generation Evaluation with Batch Centering and Tempered Word Mover Distance}

\author{Xi Chen\thanks{\;\;Work done during the internship at Google.} \\ Harvard University \\ Cambridge, MA 02138 \\\texttt{chenx@g.harvard.edu} \And Nan Ding\thanks{\;\;Corresponding author.} \\ Google Research\\  Venice, CA 90291  \And Tomer Levinboim \\ Google Research \\  Venice, CA 90291 \\
  \texttt{\{dingnan,tomerl,rsoricut\}@google.com} \And Radu Soricut \\ Google Research \\   Venice, CA 90291  }
\date{August 2020}

\begin{document}

\maketitle

\begin{abstract}
Recent advances in automatic evaluation metrics for text have shown that deep contextualized word representations, such as those generated by BERT encoders, are helpful for designing metrics that correlate well with human judgements.
At the same time, it has been argued that contextualized word representations exhibit sub-optimal statistical properties for encoding the true similarity between words or sentences. In this paper, we present two techniques for improving encoding representations for similarity metrics: a batch-mean centering strategy that improves statistical properties; and a computationally efficient tempered Word Mover Distance, for better fusion of the information in the contextualized word representations. We conduct numerical experiments that demonstrate the robustness of our techniques, reporting results over various BERT-backbone learned metrics and achieving state of the art correlation with human ratings on several benchmarks.
\end{abstract}

\section{Introduction}
Automatic evaluation metrics play an important role in comparing candidate sentences generated by machines against human references. First-generation metrics such as BLEU \cite{papineni-etal:2002} and ROUGE \cite{lin2004rouge} use predefined handcrafted rules to measure surface similarity between sentences and have no ability, or very little ability \cite{meteor}, to go beyond word surface level.
To address this problem, later work \cite{kusner2015word, zhelezniak2019correlations} utilize static embedding techniques such as word2vec \cite{mikolov-et-al:2013d} and Glove \cite{pennington-et-al:2014} to represent the words in sentences as vectors in a low-dimensional continuous space, so that word-to-word correlation can be measured by their cosine similarity.
However, static embeddings cannot capture the rich syntactic, semantic, and pragmatic aspects of word usage across sentences and paragraphs. 

Modern deep learning models based on the Transformer \cite{vaswani2017attention} utilize a multi-layered self-attention structure that encodes not only a global representation of each word (a word embedding), but also its contextualized information within the context considered.
Such contextualized word representations have yielded significant improvements on various tasks, including machine translation \cite{vaswani2017attention}, NLU tasks \cite{devlin2018bert, liu2019roberta, lan2020albert}, summarization \cite{zhang2019pegasus}, and automatic evaluation metrics~\cite{reimers2019sentence, zhang2019bertscore, zhao2019moverscore, sellam2020bleurt}.

In this paper, we investigate how to better use BERT-based contextualized embeddings in order to arrive at effective  evaluation metrics for generated text.
We formalize a unified family of text similarity metrics, which operate either at the word/token or sentence level, and show how a number of existing embedding-based similarity metrics belong to this family. 
In this context, we present a tempered Word Mover Distance (TWMD) formulation by utilizing the Sinkhorn distance~\cite{cuturi2013sinkhorn}, which adds an entropy regularizer to the objective of WMD~\cite{kusner2015word}.
Compared to WMD, our TWMD formulation allows for a more efficient optimization using the iterative Sinkhorn algorithm~\cite{cuturi2013sinkhorn}.
Although in theory the Sinkhorn algorithm may require a number of iterations to converge, we find that a single iteration is sufficient and surprisingly effective for TWMD. 

Moreover, we follow \cite{ethayarajh2019contextual} and carefully analyze the similarity between contextualized word representations along the different layers of a BERT model.
We posit three properties that multi-layered contextualized word representations should have (Section \ref{sec:centering}):
(1) zero expected similarity between random words,
(2) decreasing out-of-context self-similarity, and
(3) increasing in-context similarity between words.
As already shown by \citet{ethayarajh2019contextual}, cosine similarity between BERT word-embeddings does not satisfy some of these properties. 
To address these issues, we design and analyze several centering techniques and find one that satisfies the three properties above.
The usefulness of the centering technique and TWMD formulation is validated by our empirical studies over several well-known benchmarks, where we obtain significant numerical improvements and SoTA correlations with human ratings.

\section{Related Work}
Recent work on learned automatic evaluation metrics leverage pretrained contextualized embeddings by building on top of BERT \cite{devlin2018bert} or variant \cite{liu2019roberta} representations.

SentenceBERT~\cite{reimers2019sentence} uses cosine similarity of two mean-pooled sentence embedding from the top layer of BERT.
BERTscore \cite{zhang2019bertscore} computes the similarity of two sentences as a sum of cosine similarities between maximum-matching tokens embeddings.
MoverScore \cite{zhao2019moverscore} measures word distance using BERT embeddings and computes the Word Mover Distance (WMD) \cite{kusner2015word} from the word distribution of the system text to that of the human reference.


In the next section we propose an abstract framework of embedding-based similarity metrics and show that it contains the metrics mentioned above. We then extend this family of metrics with our own improved evaluation metric.

\section{A Family of Similarity Metrics}
We consider a family of normalized similarity metrics for both word-level and sentence-level representations parameterized by a function $C$, as follows:
\begin{align}
\mathrm{Sim}(\xb_1, \xb_2) = \frac{C(\xb_1,\xb_2)}{\sqrt{C(\xb_1,\xb_1)C(\xb_2,\xb_2)}}. \label{eq:sim_family}
\end{align}
Clearly, $\mathrm{Sim}(\xb, \xb) = 1$, and furthermore, if $C(\xb_1,\xb_2)^2 \le C(\xb_1,\xb_1)C(\xb_2,\xb_2)$, then $\mathrm{Sim}(\xb_1, \xb_2) \in [-1, 1]$.

For word similarity, $\xb$ represents a single word vector. A standard choice is defining $C(\xb_1, \xb_2) = \inner{\xb_1}{\xb_2}$, the inner product between the two vectors.
The resulting word similarity metric $\mathrm{Sim}(\xb_1, \xb_2) = \inner{\frac{\xb_1}{\|\xb_1\|}}{\frac{\xb_2}{\|\xb_2\|}}$ becomes the cosine similarity between the two word vectors. If the word vectors are pre-normalized such that $\|\xb\| = 1$, then $\mathrm{Sim}(\xb_1, \xb_2) = \inner{\xb_1}{\xb_2}$. 

For sentence similarity, we use $\Xb = \rbr{\xb^1, \xb^2, \ldots, \xb^{L}}$ to denote a $D \times L$ matrix composed by $L$ word vectors belonging to the sentence embedded in a $D$-dimensional space.

In what follows, we briefly review existing sentence similarity metrics and show that they belong to our family of similarity metrics Eq.\eqref{eq:sim_family} with different choices of $C(\Xb_1, \Xb_2)$
(with $L_1$ and $L_2$ denoting the sentence length for $\Xb_1$ and $\Xb_2$, respectively).
Note that we do not consider word re-weighting schemes (e.g. by IDF as in~\cite{zhang2019bertscore}) in this paper, as their contribution does not appear to be consistent over various tasks.
In addition, we assume that all word vectors are already pre-normalized.

\paragraph{Sentence-BERT}
Sentence-BERT~\cite{reimers2019sentence} uses the cosine-similarity between two mean-pooling sentence embeddings. This is the same as Eq.\eqref{eq:sim_family} when 
\begin{align*}
    C(\Xb_1,\Xb_2) &= \inner{\frac{1}{L_1} \sum_{i=1}^{L_1} \xb_1^i}{\frac{1}{L_2} \sum_{j=1}^{L_2} \xb_2^j} \\
    &= \frac{1}{L_1 L_2} \sum_{i=1}^{L_1} \sum_{j=1}^{L_2} \inner{\xb_1^i}{\xb_2^j}.
\end{align*}

\paragraph{Wordset-CKA}
Wordset-CKA~\cite{zhelezniak2019correlations} uses the centered kernel alignment between the two sentences represented as word sets, where
\begin{align*}
    C(\Xb_1,\Xb_2) &= \text{Tr} \rbr{\Xb_1 \Xb_1^{\top} \Xb_2 \Xb_2^{\top}} \\
    &= \sum_{i=1}^{L_1} \sum_{j=1}^{L_2} \inner{\xb_1^i}{\xb_2^j}^2.
\end{align*}
Here we assume each word embedding $\xb$ is pre-centered by the mean of its own dimensions.
We refer to this centering method as \emph{dimension-mean centering}. 

\paragraph{MoverScore}
MoverScore~\cite{zhao2019moverscore} measures the sentence similarity using the Word Mover Distance \cite{kusner2015word} from the word distribution of the hypothesis to that of the gold reference: 
\begin{align}
     C(\Xb_1,\Xb_2)
    &= \max_{\pi}\sum_{i=1}^{L_1} \sum_{j=1}^{L_2}  \pi_{ij} \inner{\xb_1^i}{\xb_2^j} \nonumber\\
    \text{s.t.} \;\; \sum_{i=1}^{L_1} \pi_{ij} &= \frac{1}{L_2}, \;\; \sum_{j=1}^{L_2} \pi_{ij} = \frac{1}{L_1}. \label{eq:moverscore}
\end{align}
The original MoverScore does not normalize $C(\Xb_1,\Xb_2)$ by $\sqrt{C(\Xb_1,\Xb_1)C(\Xb_2,\Xb_2)}$. In practice, we find the performance to be similar with or without such normalization.

\paragraph{BERTscore}
BERTscore~\cite{zhang2019bertscore} introduces three metrics corresponding to recall, precision, and F1 score.
We focus the discussion here on BERTscore-Recall, as it performs most consistently across all tasks (see discussions of the precision and F1 scores in Appendix \ref{sec:f1}).
BERTscore-Recall uses the sum of cosine similarities between maximum-matching tokens embeddings: 
\begin{align}
    C(\Xb_1,\Xb_2)
    &= \frac{1}{L_1} \sum_{i=1}^{L_1} \max_{j=1\ldots L_2} \inner{\xb_1^i}{\xb_2^j}. \label{eq:bertscore}
\end{align}
For BERTscore, since the words are pre-normalized, we have $C(\Xb_1,\Xb_1) = C(\Xb_2,\Xb_2) = 1$ and therefore $\mathrm{Sim}(\Xb_1,\Xb_2) = C(\Xb_1,\Xb_2)$. 

Note that BERTscore is closely related to MoverScore, since Eq.\eqref{eq:bertscore} is the solution of the Relaxed-WMD~\cite{kusner2015word}: 
\begin{align}
     C(\Xb_1,\Xb_2)
    &= \max_{\pi}\sum_{i=1}^{L_1} \sum_{j=1}^{L_2}  \pi_{ij} \inner{\xb_1^i}{\xb_2^j} \nonumber\\
    \text{s.t.} \;\; \sum_{j=1}^{L_2} \pi_{ij} &= \frac{1}{L_1}.
    \label{eq:bertscoreopt}
\end{align}
which is the same as Eq.\eqref{eq:moverscore} but without the first constraint.

\section{Tempered Word Mover Distance}
Word Mover Distance~\cite{kusner2015word} used in MoverScore~\cite{zhao2019moverscore} is rooted in the classical optimal transport distance for probability measures and histograms of features.
Despite its excellent performance and intuitive formulation, its computation involves a linear programming solver whose cost scales as $O(L^3 \log L)$ and becomes prohibitive for long sentences or documents with more than a few hundreds of words/tokens.
For this reason, \cite{kusner2015word} proposed a Relaxed-WMD (RWMD) with only one constraint (see Eq.\eqref{eq:bertscoreopt}), which can be evaluated in $O(L^2)$.
However, RWMD uses the closest distance without considering there may be multiple words transforming to single words.

Inspired by the Sinkhorn distance~\cite{cuturi2013sinkhorn} which smooths the classic optimal transport problem with an entropic regularization term, we introduce the following formulation, which we refer to as tempered-WMD (TWMD): 
\begin{align}
     &\max_{\pi}\sum_{i=1}^{L_1} \sum_{j=1}^{L_2}  \pi_{ij} \inner{\xb_1^i}{\xb_2^j} - T \sum_{i=1}^{L_1} \sum_{j=1}^{L_2}  \pi_{ij} \log \pi_{ij} \nonumber\\
    &\text{s.t.} \;\; \sum_{i=1}^{L_1} \pi_{ij} = \frac{1}{L_2}, \;\; \sum_{j=1}^{L_2} \pi_{ij} = \frac{1}{L_1}. \label{eq:sinkhorn_distance}
\end{align}
The temperature parameter $T \ge 0$ determines the trade-off between the two terms. When $T = 0$, Eq.\eqref{eq:sinkhorn_distance} reduce to the original WMD as in Eq.\eqref{eq:moverscore}. When $T$ is larger, \eqref{eq:sinkhorn_distance} encourages more homogeneous distributions.

The added entropy term makes Eq.\eqref{eq:sinkhorn_distance} a strictly concave problem, which can be solved using a matrix scaling algorithm with a linear convergence rate.
For example, the Sinkhorn algorithm~\cite{cuturi2013sinkhorn} uses the initial condition $\pi_{ij}^0 = \exp\rbr{-\frac{1}{T}\inner{\xb_1^i}{\xb_2^j}}$ and alternates between
\begin{align}
\xi_{ij}^{t}=\frac{\pi_{ij}^{t-1}}{L_2\sum_i \pi_{ij}^{t-1}}, \;\;
\pi_{ij}^{t}=\frac{\xi_{ij}^{t}}{L_1\sum_j \xi_{ij}^{t}}. \label{eq:sinkhorn}
\end{align}
The computational cost for each iteration is $O(L^2)$, which is more efficient than to that of WMD. 
Although in theory this iterative algorithm may require a few of iterations to converge, our experiments show that a single iteration (i.e., $t=1$) is sufficient and surprisingly effective.

Similarly, a tempered-RWMD (TRWMD) can be obtained by adding an entropy term to Eq.\eqref{eq:bertscoreopt}:
\begin{align*}
     &\max_{\pi}\sum_{i=1}^{L_1} \sum_{j=1}^{L_2}  \pi_{ij} \inner{\xb_1^i}{\xb_2^j} - T \sum_{i=1}^{L_1} \sum_{j=1}^{L_2}  \pi_{ij} \log \pi_{ij} \nonumber\\
    &\text{s.t.} \;\; \;\; \sum_{j=1}^{L_2} \pi_{ij} = \frac{1}{L_1}.
\end{align*}
By taking the derivative of the Lagrangian of the above objective, the following closed-form solution is obtained:
\begin{align*}
     \pi^*_{ij} =& \frac{1}{L_1} \text{softmax}_j \rbr{\frac{1}{T}\inner{\xb_1^i}{\xb_2^j}}. 
\end{align*}
Plugging in the optimal $\pi^*_{ij}$ back into the objective yields the following metric:
\begin{align}
     & C(\Xb_1, \Xb_2) = \nonumber\\
     =& \frac{T}{L_1} \sum_{i=1}^{L_1}  \log \rbr{\sum_{j=1}^{L_2} \exp\rbr{\frac{1}{T}\inner{\xb_1^i}{\xb_2^j}}}.\label{eq:temp_bertscore}
\end{align}
We note that as $T \to 0$, $T\log \sum_j \exp(f_j/T) \to \max_j(f_j)$, and therefore Eq.\eqref{eq:temp_bertscore} reduces to Eq.\eqref{eq:bertscore}.

\section{Centered Word Vectors}
\label{sec:centering}
\begin{figure}[!h]
    \centering
    \includegraphics[width=0.49\textwidth]{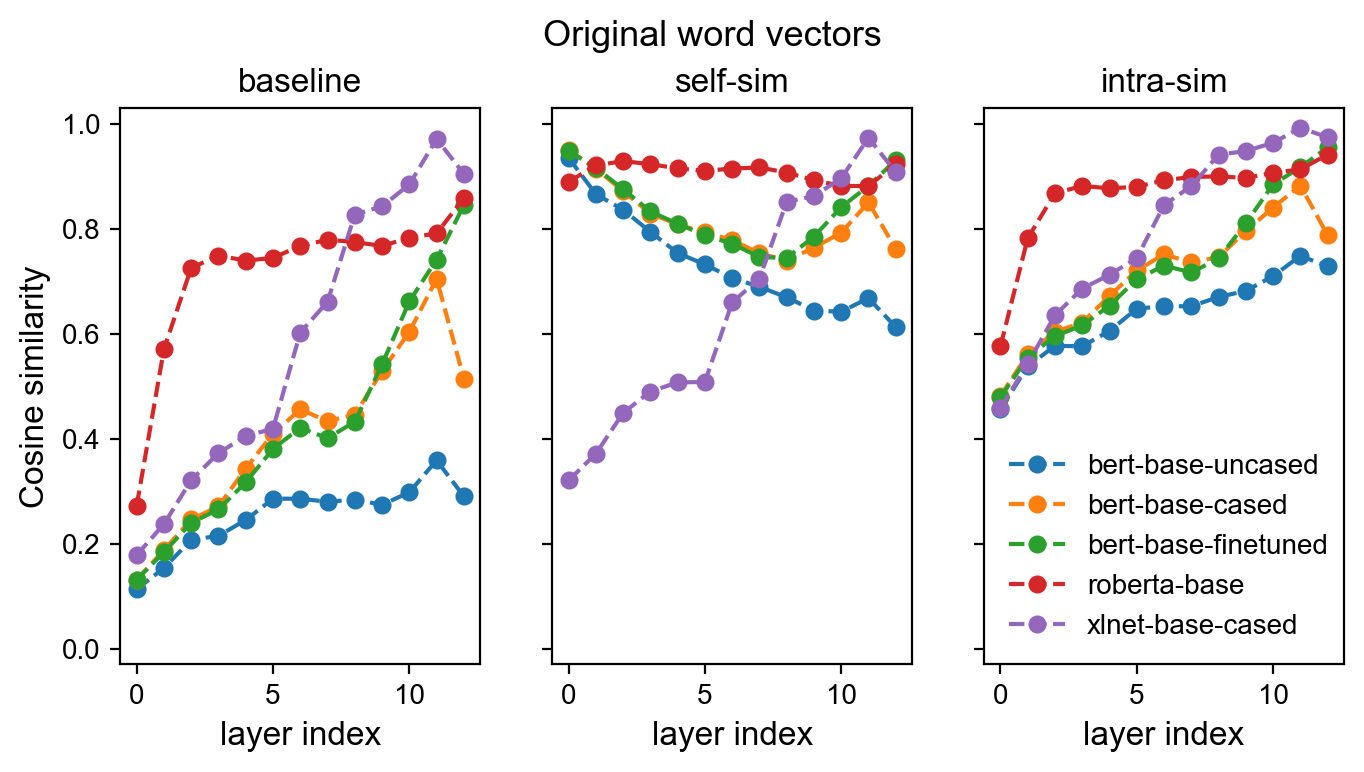}
    \caption{Cosine similarity between two random words (baseline), same words in two different sentences (self-similarity) and two random words in the same sentence (intra-similarity) for five base models, using the original layer representation of words.}
    \label{fig:org_wv}
\end{figure}

\citet{ethayarajh2019contextual} reports that representations obtained by deep models such as BERT exhibit high cosine similarity between any two random words in a corpus, especially at higher layers. They attribute this phenomenon to a highly anisotropic distribution of the word vectors, and further argue that such high similarity represents a bias that blurs the true similarity relationship between word (and sentence) representations and hampers performance in NLP tasks \cite{mu2018allbutthetop}.
We reproduce here the main results of \cite{ethayarajh2019contextual}, including the cosine similarity between two random words (baseline), same words in two different sentences (self-similarity) and two random words in the same sentence (intra-similarity).
Figure \ref{fig:org_wv} shows these results for several BERT and BERT-like models.
As the leftmost figure shows, most of these models indeed have a high baseline similarity that quickly increases with layer depth.
\citet{ethayarajh2019contextual} proposes to mitigate this bias by subtracting the baseline similarity from the self-similarity and intra-similarity values (per layer). However, the mathematical and statistical meaning of this solution remains unclear. 

In this context, we posit the following three properties that are desirable for word vector representations in context:
\begin{enumerate}
    \item Zero expected similarity: The word similarity between two random word vectors in the corpus is approximately zero, which indicates random words are unrelated.
    \item Decreasing self-similarity: The word similarity between representations of the same word taken from different sentences decreases in higher layers, as each representation encodes more contextual information about its respective sentence.
    \item Increasing intra-similarity: The word similarity between different words within the same sentence increases in higher layers, as the words encode more common information about the sentence.
\end{enumerate}
Besides their intuitive appeal, our empirical results (in Section 6) do validate that word representations that obey these properties result in higher performance with respect to modeling similarity.

\begin{figure}[!h]
    \centering
    \includegraphics[width=0.49\textwidth]{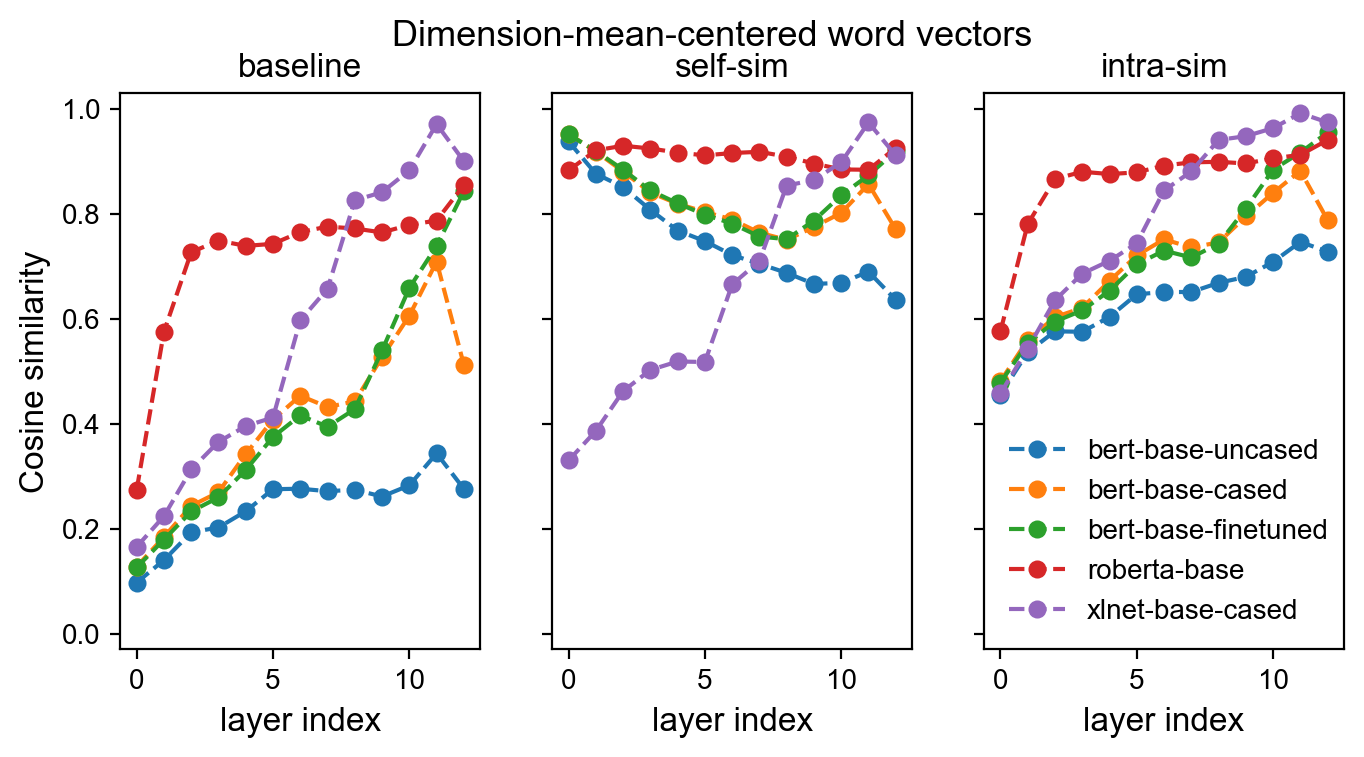}
    \includegraphics[width=0.49\textwidth]{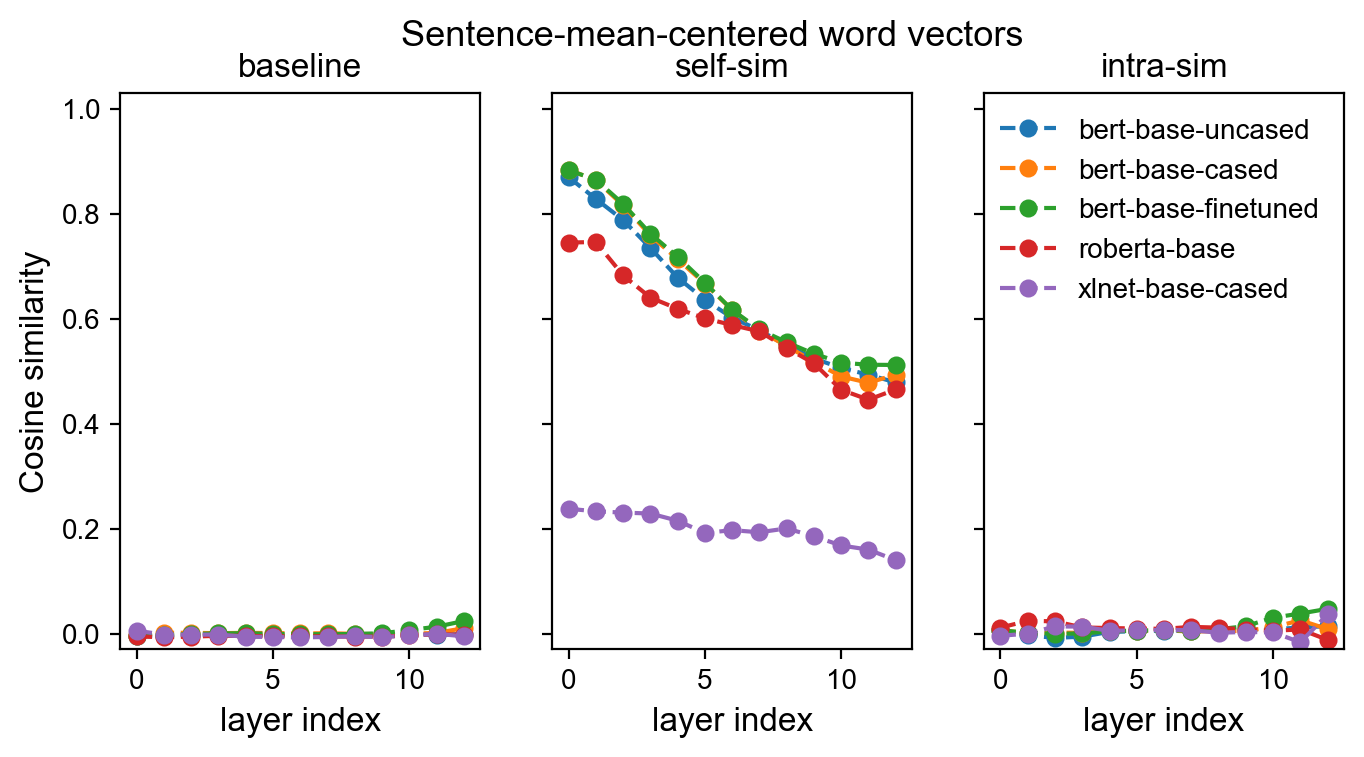}
    \includegraphics[width=0.49\textwidth]{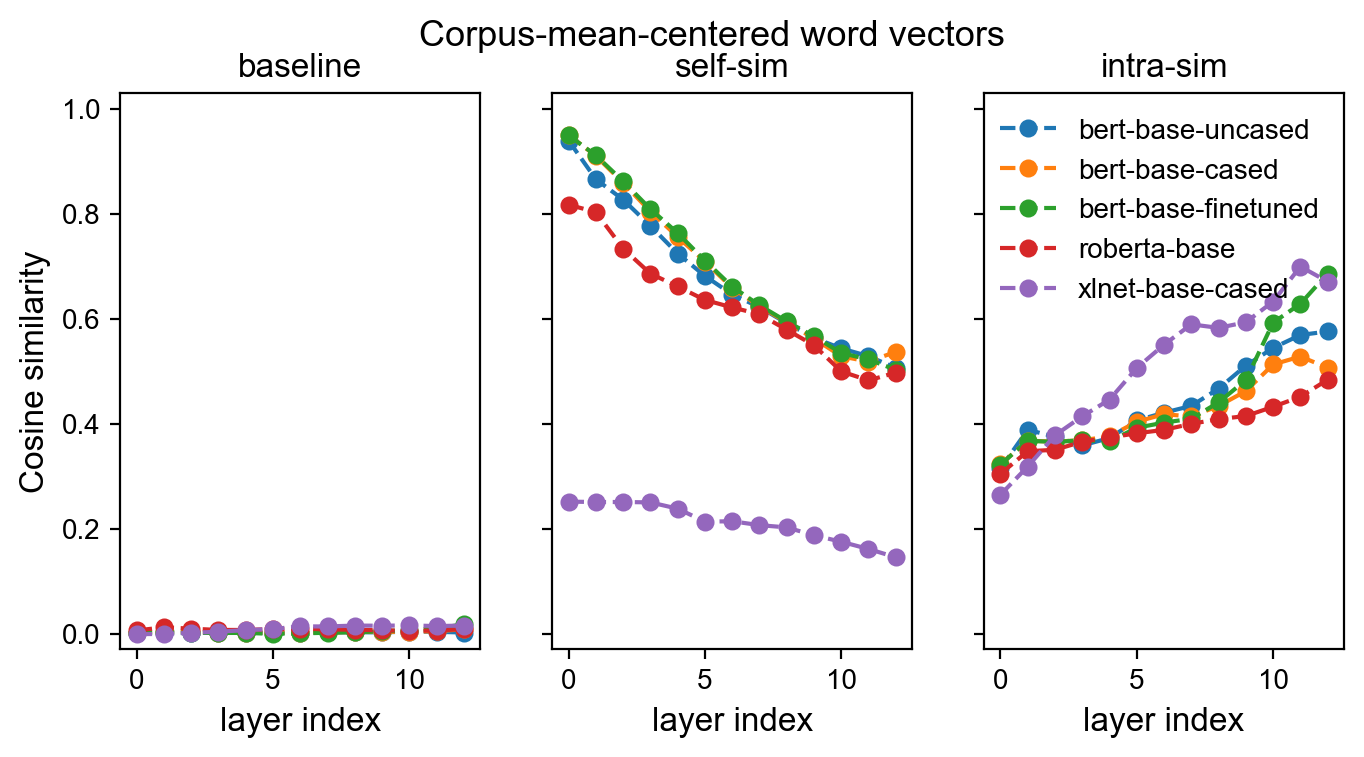}
    \caption{Comparison of the three centering approaches. Dimension mean centering has very little effect. Sentence mean centering removes too much common sentence information. Corpus mean centering shows the correct word contextualization.}
    \label{fig:corpus_center}
\end{figure}

Since the original word representations does not satisfy these three properties, we explore three methods for centering the word vectors distribution. Consider a corpus $\mathcal{C}$ containing $M$ sentences $\{s_i\}$, each of length $N_i$. Each word vector is $D$-dimensional, $\wb_{i,j} = [w_{i,j}^{(1)},...,w_{i,j}^{(D)}]$.
We propose three candidate word distribution centering approaches: 
\begin{itemize}
\item Dimension mean centering: centering a word by subtracting the mean of the dimensions within each word vector,
$$\vb_{i,j} = \wb_{i,j}-\frac{1}{D}\sum_{l=1}^D w_{i,j}^{(l)}.$$
The second term on the RHS is a scalar, which broadcasts to all dimensions of $\wb_{i,j}$.
\item Sentence mean centering: centering a word by subtracting the mean of the words within the corresponding sentence,
$$\vb_{i,j} = \wb_{i,j}- \frac{1}{N}\sum_{k=1}^N \wb_{i,k}.$$
\item Corpus mean centering: centering a word by subtracting the mean of the words in the entire corpus,
$$\vb_{i,j} = \wb_{i,j}-\frac{1}{\sum_i N_i}\sum_{i=1}^M\sum_{k=1}^{N_i} \wb_{i,k}.$$
\end{itemize}
We compare these three centering approaches in Figure \ref{fig:corpus_center}.
Due to the layer norm operation in the BERT models, the dimension mean is a small constant that has little effect after subtraction, and therefore it fails on properties 1 and 2 above.
The sentence mean centering achieves approximately zero baseline (property 1), but it also reduces the intra-sim to approximately zero (failing property 3). This indicates the subtraction of sentence mean removes the common knowledge of the words about the sentence, which can have a detrimental effect on modeling similarity.
Lastly, corpus-mean centering fulfills all three properties above (Fig.~\ref{fig:corpus_center}, bottom row). In this context, we note that, after applying corpus mean centering,  cosine-similarity function is reduced to Pearson's correlation.

Since the computational cost of corpus mean centering can be prohibitive for a large dataset, we consider a batch-mean centering approach, which would be especially useful for fine-tuning tasks. In practice, we find that the values obtained from batch-mean--centered word vectors are very close to those of corpus-mean--centered word vectors. Therefore and henceforth, we use batch-mean centering to approximate the effect of corpus-mean centering. 

Finally, it is worth noting that corpus (batch)-mean centering has recently been applied in normalizing multilingual representations~\cite{libovicky2019language,zhao2020inducing}. However, we are the first to demonstrate its superiority over various other centering methods in single-language by analyzing the inter-layer representation similarities. 

\section{Experiments}
In order to demonstrate the effectiveness of our newly proposed approaches, we conduct extensive numerical experiments based on two commonly-used benchmarks: Semantic Textual Similarity (STS), and WMT 17-18 metrics shared task.
Our experiments are designed to answer the following questions:
(1) Are corpus (batch) centered word vectors better than other centered and un-centered word vectors, across different sentence similarity metrics?
(2) How do tempered WMD and RWMD compare to their family-relatives MoverScore and BERTscore?
(3) How do the temperature hyperparameter and the Sinkhorn iterations affect the performance, and how sensitive are they?

To show that our results are consistent across different BERT variants, we analyze our similarity metrics over four backbone models: bert-base-uncased, bert-large-uncased, roberta-base and roberta-large, all obtained from the Huggingface\footnote{\url{https://huggingface.co/models}} Transformers package. 
\cite{zhang2019bertscore} found that the better layers for evaluation metric are usually not the top layer, since the top one is greatly impacted by the pretraining task. In particular, \cite{zhang2019bertscore}  perform an an extensive layer sweep analysis and report that the better layers were always around Layer-10 for the base models, and Layer-19 for the large models. Therefore, in our experiments, we used Layer-10 for all base models, and Layer-19 for all large models. We also present the results of evaluation metrics using different layers in the Appendix \ref{sec:layer} and confirm that our main conclusion is not affected by the choice of layer. 

\begin{table*}[!h]
\centering
\scriptsize
\caption{Experimental results for various metrics on STS 12-16 datasets (averaged) with BERT/Roberta pretrained checkpoints. The correlations are Pearson (left) and Spearman's rank (right). }
\label{table:sts}
\begin{tabular}{c  c c c c c}
\hline
\textbf{Metric} &\makecell[c]{\bf bert-base-uncased\\$r\ /\ \rho$} &\makecell[c]{\bf bert-large-uncased\\$r\ /\ \rho$} &\makecell[c]{\bf roberta-base\\$r\ /\ \rho$} &\makecell[c]{\bf roberta-large\\$r\ /\ \rho$} \\\hline
\makecell[c]{SBERT} & 58.7 / 58.9 & 56.9 / 57.3 & 58.0 / 59.6 & 58.5 / 60.2 \\
\makecell[c]{SBERT-batch} & \textbf{63.8 / 62.8} & \textbf{62.8 / 62.3} & \textbf{65.9 / 65.1} & \textbf{67.1 / 66.3} \\
\makecell[c]{SBERT-dim} & 58.7 / 58.9 & 56.9 / 57.3 & 58.0 / 59.6 & 58.5 / 60.2 \\\hline
\makecell[c]{CKA} & 59.8 / 59.5 & 58.7 / 58.9 & 58.6 / 59.9 & 59.1 / 60.4 \\
\makecell[c]{CKA-batch} & \bf 60.3 / 61.1 & 58.9 / 60.0 & \bf 61.1 / 61.5 & \bf 62.3 / 62.5 \\
\makecell[c]{CKA-sent} & 58.6 / 59.8 & \bf 59.1 / 60.5 & 58.7 / 59.2 & 60.6 / 61.0 \\
\makecell[c]{CKA-dim} & 59.8 / 59.5 & 58.7 / 58.9 & 58.6 / 59.9 & 59.1 / 60.4 \\\hline
\makecell[c]{MoverScore} & 56.3 / 58.2 & 54.4 / 56.7 & 54.8 / 56.2 & 54.5 / 56.0 \\
\makecell[c]{MoverScore-batch} & \bf 58.0 / 60.1 & \bf 56.2 / 58.6 & \bf 57.2 / 59.0 & \bf 57.7 / 59.3 \\
\makecell[c]{MoverScore-sent} & 54.2 / 57.4 & 54.9 / 58.3 & 54.1 / 56.5 & 55.9 / 58.1 \\
\makecell[c]{MoverScore-dim} & 56.3 / 58.2 & 54.4 / 56.7 & 54.8 / 56.2 & 54.5 / 56.0 \\\hline
\makecell[c]{BERTscore} & 59.3 / 59.0 & 57.7 / 57.8 & 57.3 / 57.2 & 57.0 / 57.1 \\
\makecell[c]{BERTscore-batch} & \bf 61.1 / 60.9 & \bf 59.6 / 59.7 & \bf 60.6 / 60.6 & \bf 61.5 / 61.4 \\
\makecell[c]{BERTscore-sent} & 57.3 / 57.6 & 58.1 / 58.6 & 56.8 / 57.2 & 59.0 / 59.2 \\
\makecell[c]{BERTscore-dim} & 59.3 / 59.0 & 57.7 / 57.8 & 57.3 / 57.2 & 57.0 / 57.1 \\
\hline
\end{tabular}
\end{table*}

\subsection{Semantic Textual Similarity (STS)}
The STS benchmark \cite{agirre2016semeval} contains sentence pairs and human evaluated scores between 0 and 5 for each pair, with higher scores indicating higher semantic relatedness or similarity for the pair.
From 2012 to 2016, it contains 3108, 1500, 3750, 3000, and 1186 records, respectively.

We answer the first question by comparing batch-centered word vectors with other centered and un-centered word vectors using several sentence similarity metrics, including Sentence-BERT, Wordset-CKA, BERTscore and MoverScore. 
 
The results on STS 12-16 for various metrics are shown in Table \ref{table:sts}.
In general, for all four models (per column: base and large version of BERT and RoBERTa), batch centering gets higher Pearson and Spearman's correlation of sentence-mean cosine similarity (SBERT) and BERTscore.
Dimension-mean centering has very little effect on performance, while sentence-mean improves performance for a few methods.
Since Sentence-BERT uses the mean-pooling of the sentence (which would become zero after sentence mean centering), we exclude sentence-mean centering from Sentence-BERT.
Overall, batch-mean centering brings an averaged +3.41 / +3.02 improvement, and sentence-mean centering brings an averaged -0.02 / +0.55 on Pearson and Spearman coefficients, across different metrics and models.

\begin{table*}[!ht]
\scriptsize
\centering
\caption{Correlation with human scores on the WMT17 Metrics Shared Task. `-b' stands for batch centering of word vectors. }
\label{table:wmt17}
\begin{tabular}{c c c c c c c c c}
\hline
\bf Metric & \bf \makecell[c]{\bf cs-en\\$\tau\ /\ r$} & \bf \makecell[c]{\bf de-en\\$\tau\ /\ r$} & \bf \makecell[c]{\bf fi-en\\$\tau\ /\ r$} & \bf \makecell[c]{\bf lv-en\\$\tau\ /\ r$} & \bf \makecell[c]{\bf ru-en\\$\tau\ /\ r$} & \bf \bf \makecell[c]{\bf tr-en\\$\tau\ /\ r$} & \bf \makecell[c]{\bf zh-en\\$\tau\ /\ r$} & \bf \makecell[c]{\bf Avg.\\$\tau\ /\ r$}\\\hline
\emph{roberta-base}\\\hline
\makecell[c]{SBERT} & 45.1 / 60.0 & 44.6 / 58.3 & 58.4 / 69.6 & 42.9 / 60.6 & 45.8 / 63.1 & 46.3 / 52.9 & 46.0 / 62.0 & 47.0 / 60.9\\
\makecell[c]{SBERT-b} & 45.2 / 63.4 & 45.8 / 64.1 & 56.8 / 74.6 & 45.1 / 64.9 & 44.9 / 64.0 & 47.8 / 63.4 & 45.4 / 66.1 & 47.3 / 65.8\\
\makecell[c]{CKA} & 45.0 / 60.5 & 44.8 / 58.8 & 58.3 / 70.5 & 42.8 / 61.0 & 45.9 / 63.4 & 46.3 / 53.9 & 46.1 / 62.4 & 47.0 / 61.5\\
\makecell[c]{CKA-b} & 48.8 / 68.4 & 49.1 / 69.1 & 61.3 / 81.3 & 48.5 / 69.6 & 49.6 / 69.6 & 52.1 / 71.7 & 49.6 / 70.8 & 51.3 / 71.5\\
\makecell[c]{MoverScore} & 48.5 / 66.0 & 47.1 / 65.9 & 61.6 / 80.9 & 48.9 / 68.2 & 51.6 / 69.8 & 53.8 / 74.2 & 53.4 / 74.0 & 52.1 / 71.3\\
\makecell[c]{MoverScore-b} & 47.9 / 66.3 & 47.3 / 66.1 & 61.6 / 81.2 & 48.6 / 68.6 & 51.4 / 69.8 & 54.3 / 74.9 & 52.2 / 72.8 & 51.9 / 71.3\\
\makecell[c]{BERTscore} & 47.4 / 64.7 & 48.0 / 66.9 & 61.9 / 79.9 & 49.7 / 69.6 & 50.8 / 69.5 & 53.4 / 71.3 & 50.8 / 71.7 & 51.7 / 70.5\\
\makecell[c]{BERTscore-b} & 47.5 / 66.4 & 48.8 / 68.7 & 61.7 / 81.3 & 49.9 / 70.6 & 50.7 / 69.8 & 53.8 / 73.2 & 49.1 / 70.1 & 51.6 / 71.5\\
\makecell[c]{TWMD} & 48.3 / 65.8 & 49.6 / 68.8 & 62.5 / 81.2 & 51.3 / 70.5 & 52.1 / 71.2 & 54.6 / 73.8 & \bf 54.7 / 75.5 & 53.3 / 72.3\\
\makecell[c]{TWMD-b} & \bf 50.0 / 68.5 & \bf 51.5 / 70.8 & \bf 63.0 / 82.8 & \bf 51.9 / 72.3 & \bf 53.5 / 73.2 & \bf 56.6 / 77.0 & 54.0 / 75.0 & \bf 54.4 / 74.3\\
\makecell[c]{TRWMD} & 47.4 / 64.9 & 47.9 / 67.0 & 61.8 / 80.1 & 49.5 / 69.3 & 50.9 / 69.5 & 53.4 / 71.8 & 50.7 / 71.7 & 51.7 / 70.7\\
\makecell[c]{TRWMD-b} & 48.5 / 66.8 & 49.0 / 68.5 & 61.1 / 81.3 & 49.5 / 69.3 & 51.4 / 69.8 & 54.3 / 74.7 & 50.2 / 70.8 & 52.0 / 71.6\\

\hline

\emph{roberta-large}\\\hline
\makecell[c]{SBERT} & 50.9 / 67.2 & 53.1 / 70.8 & 61.3 / 73.6 & 51.6 / 70.5 & 51.4 / 69.0 & 52.4 / 61.4 & 51.9 / 68.0 & 53.2 / 68.6\\
\makecell[c]{SBERT-b} & 47.6 / 66.9 & 50.7 / 69.5 & 56.8 / 74.1 & 47.9 / 67.8 & 47.3 / 66.4 & 48.5 / 65.2 & 47.6 / 67.5 & 49.5 / 68.2\\
\makecell[c]{CKA} & 51.4 / 68.7 & 53.4 / 71.3 & 61.5 / 74.5 & 51.8 / 71.1 & 51.8 / 69.3 & 52.7 / 62.7 & 52.1 / 68.8 & 53.5 / 69.5\\
\makecell[c]{CKA-b} & 51.6 / 72.3 & 54.4 / 74.2 & 61.8 / 81.6 & 52.5 / 73.7 & 53.2 / 73.0 & 53.6 / 73.8 & 52.7 / 73.5 & 54.3 / 74.6\\
\makecell[c]{MoverScore} & 51.6 / 68.8 & 53.9 / 71.8 & 62.0 / 81.1 & 53.4 / 71.7 & 54.5 / 71.8 & 56.3 / 76.2 & 56.3 / 76.1 & 55.5 / 73.9\\
\makecell[c]{MoverScore-b} & 51.2 / 69.6 & 53.2 / 71.7 & 63.1 / 82.1 & 53.3 / 72.7 & 54.5 / 72.8 & 56.8 / 76.9 & 55.1 / 75.4 & 55.3 / 74.5\\
\makecell[c]{BERTscore} & 50.9 / 66.9 & 53.4 / 72.3 & 61.7 / 79.6 & 53.5 / 71.6 & 53.8 / 71.5 & 54.8 / 71.7 & 53.9 / 74.4 & 54.6 / 72.6\\
\makecell[c]{BERTscore-b} & 51.7 / 71.2 & 53.9 / 74.1 & 63.6 / 82.5 & 54.8 / \bf 75.1 & 54.8 / 73.7 & 55.6 / 75.0 & 52.7 / 73.6 & 55.3 / 75.0\\
\makecell[c]{TWMD} & 52.3 / 69.1 & 55.7 / 74.4 & 63.1 / 81.5 & 54.1 / 72.6 & 56.0 / 74.1 & 55.7 / 74.5 & \bf 57.5 / 77.7 & 56.3 / 74.9\\
\makecell[c]{TWMD-b} & \bf 53.9 / 73.3 & \bf 56.4 / 75.9 & \bf 64.4 / 83.5 & \bf 55.2 / 75.1 & \bf 56.9 / 76.2 & \bf 57.9 / 78.1 & 56.8 / 77.4 & \bf 57.4 / 77.1\\
\makecell[c]{TRWMD} & 50.8 / 67.3 & 53.3 / 72.1 & 61.5 / 79.7 & 53.1 / 71.3 & 54.0 / 71.5 & 54.5 / 72.0 & 54.0 / 74.3 & 54.5 / 72.6\\
\makecell[c]{TRWMD-b} & 52.5 / 71.2 & 53.9 / 73.4 & 62.7 / 82.0 & 53.8 / 73.4 & 54.8 / 72.8 & 55.7 / 76.1 & 53.4 / 74.1 & 55.3 / 74.7\\
\hline

\end{tabular}
\end{table*}

\begin{table*}[!h]
\scriptsize
\centering
\caption{Correlation with human scores on the WMT18 Metrics Shared Task. `-b' stands for batch centering of word vectors. }
\label{table:wmt18}
\begin{tabular}{c c c c c c c c c}
\hline
\bf Metric & \bf \makecell[c]{\bf cs-en\\$\tau\ /\ r$} & \bf \makecell[c]{\bf zh-en\\$\tau\ /\ r$} & \bf \makecell[c]{\bf ru-en\\$\tau\ /\ r$} & \bf \makecell[c]{\bf fi-en\\$\tau\ /\ r$} & \bf \makecell[c]{\bf tr-en\\$\tau\ /\ r$} & \bf \bf \makecell[c]{\bf et-en\\$\tau\ /\ r$} & \bf \makecell[c]{\bf de-en\\$\tau\ /\ r$} & \bf \makecell[c]{\bf Avg.\\$\tau\ /\ r$}\\\hline

\emph{roberta-base}\\\hline
\makecell[c]{SBERT} & 26.2 / 34.2 & 22.9 / 29.0 & 23.5 / 34.2 & 23.1 / 32.4 & 25.3 / 34.6 & 30.3 / 42.3 & 35.6 / 48.5 & 26.7 / 36.5\\
\makecell[c]{SBERT-b} & 28.7 / 40.6 & 25.8 / 35.6 & 26.5 / 38.4 & 25.1 / 36.6 & 28.7 / 39.8 & 33.4 / 46.9 & 38.4 / 54.1 & 29.5 / 41.6\\
\makecell[c]{CKA} & 26.2 / 34.6 & 23.0 / 29.5 & 23.5 / 34.5 & 23.2 / 32.7 & 25.4 / 35.1 & 30.4 / 42.6 & 35.8 / 48.9 & 26.8 / 36.9\\
\makecell[c]{CKA-b} & 28.9 / 41.6 & 26.4 / 37.4 & 27.2 / 39.9 & 25.7 / 37.8 & 30.0 / 42.6 & 34.2 / 49.3 & 39.0 / 55.7 & 30.2 / 43.5\\
\makecell[c]{MoverScore} & 28.5 / 40.6 & 28.2 / 37.8 & 27.9 / 39.4 & 25.1 / 35.8 & 31.3 / 43.4 & 34.3 / 48.5 & 38.9 / 54.2 & 30.5 / 42.8\\
\makecell[c]{MoverScore-b} & 28.7 / 40.6 & 28.1 / 37.7 & 27.8 / 39.2 & 25.2 / 36.0 & 31.3 / 43.2 & 34.4 / 48.6 & 38.9 / 54.2 & 30.5 / 42.8\\
\makecell[c]{BERTscore} & 27.6 / 39.5 & 27.2 / 37.5 & 27.6 / 39.4 & 24.9 / 36.0 & 31.1 / 43.3 & 34.4 / 48.6 & 39.3 / 56.1 & 30.3 / 42.9\\
\makecell[c]{BERTscore-b} & 27.8 / 39.9 & 26.9 / 37.5 & 27.5 / 39.3 & 25.2 / 37.1 & 31.1 / 43.4 & 34.4 / 49.0 & 39.5 / 56.5 & 30.4 / 43.2\\
\makecell[c]{TWMD} & 28.7 / 40.9 & 28.2 / 38.4 & 28.1 / 39.9 & 25.5 / 36.7 & 31.7 / 43.9 & 34.9 / 49.3 & 39.9 / 56.6 & 31.0 / 43.7\\
\makecell[c]{TWMD-b} & \bf 29.5 / 42.0 & \bf 28.4 / 38.9 & \bf 28.7 / 40.6 & \bf 26.4 / 38.2 & \bf 32.3 / 44.5 & \bf 35.4 / 50.2 & \bf 40.4 / 57.4 & \bf 31.6 / 44.6\\
\makecell[c]{TRWMD} & 27.6 / 39.4 & 27.3 / 37.6 & 27.6 / 39.3 & 24.9 / 35.9 & 31.1 / 43.3 & 34.4 / 48.5 & 39.4 / 56.2 & 30.3 / 42.9\\
\makecell[c]{TRWMD-b} & 28.1 / 39.9 & 27.6 / 37.6 & 27.9 / 39.2 & 25.3 / 36.6 & 31.4 / 43.2 & 34.6 / 49.1 & 39.9 / 56.7 & 30.7 / 43.2\\
\hline
\emph{roberta-large}\\\hline
\makecell[c]{SBERT} & 29.0 / 40.4 & 24.9 / 33.2 & 26.6 / 38.2 & 26.2 / 37.6 & 28.4 / 39.3 & 33.9 / 44.1 & 39.0 / 54.3 & 29.7 / 41.0\\
\makecell[c]{SBERT-b} & 30.4 / 42.6 & 26.6 / 35.8 & 27.9 / 38.9 & 27.0 / 38.7 & 29.9 / 41.1 & 35.0 / 47.1 & 40.1 / 55.9 & 31.0 / 42.9\\
\makecell[c]{CKA} & 29.2 / 40.8 & 25.0 / 33.6 & 26.7 / 38.6 & 26.3 / 37.8 & 28.5 / 39.7 & 34.0 / 44.7 & 39.1 / 54.7 & 29.7 / 41.4\\
\makecell[c]{CKA-b} & 30.4 / 43.9 & 27.0 / 37.8 & 28.2 / 41.0 & 27.0 / 39.5 & 30.5 / 43.5 & 35.5 / 50.6 & 40.4 / 57.6 & 31.3 / 44.8\\
\makecell[c]{MoverScore} & 29.9 / 41.8 & 28.7 / 38.0 & 29.2 / 40.2 & 26.5 / 37.2 & 31.9 / 43.6 & 35.9 / 49.9 & 40.8 / 56.2 & 31.9 / 43.8\\
\makecell[c]{MoverScore-b} & 30.0 / 42.1 & 28.7 / 38.1 & 28.9 / 40.0 & 26.5 / 37.5 & 31.7 / 43.5 & 35.6 / 49.7 & 40.4 / 55.6 & 31.7 / 43.8\\
\makecell[c]{BERTscore} & 29.4 / 41.5 & 27.9 / 37.9 & 28.9 / 40.3 & 26.0 / 36.6 & 31.6 / 43.6 & 35.9 / 49.6 & 41.1 / 58.4 & 31.6 / 44.0\\
\makecell[c]{BERTscore-b} & 29.7 / 42.6 & 27.6 / 38.2 & 28.9 / 40.9 & 26.4 / 38.4 & 31.6 / 44.2 & 35.9 / 50.1 & 41.0 / 58.5 & 31.6 / 44.7\\
\makecell[c]{TWMD} & 30.5 / 42.9 & \textbf{28.9} / 39.0 & 29.5 / 40.9 & 27.2 / 38.1 & 32.4 / 44.4 & 36.5 / 50.5 & \textbf{41.8} / 58.9 & 32.4 / 45.0\\
\makecell[c]{TWMD-b} & \bf 31.1 / 44.2 & \bf 28.9 / 39.5 & \bf 29.7 / 41.8 & \bf 27.6 / 39.6 & \bf 32.6 / 45.1 & \bf 36.7 / 51.3 & \bf 41.8 / 59.0 & \bf 32.7 / 45.8\\
\makecell[c]{TRWMD} & 29.2 / 41.4 & 28.0 / 37.8 & 28.9 / 40.2 & 25.9 / 36.5 & 31.6 / 43.5 & 35.9 / 49.5 & 41.1 / 58.3 & 31.5 / 43.9\\
\makecell[c]{TRWMD-b} & 29.8 / 42.3 & 28.1 / 38.1 & 29.2 / 40.6 & 26.3 / 37.7 & 31.8 / 43.8 & 36.0 / 50.2 & 41.3 / 58.4 & 31.8 / 44.4\\
\hline
\end{tabular}
\end{table*}

\subsection{WMT metrics shared task}
The WMT metrics shared task is an annual competition for comparing translation metrics against human assessments on machine-translated sentences. We use years 2017 and 2018 of the official WMT test set for evaluation. The 2017 test data includes 3,920 pairs of sentences from the news domain (including a system generated sentence and a groundtruth sentence by human) with human ratings. Similarly, the 2018 test data includes 138,188 pairs of sentences with human ratings but is reported to be much noisier~\cite{sellam2020bleurt}. 

\subsubsection*{Evaluation metrics without fine-tuning} 
We compare the Tempered WMD (TWMD) and TRWMD with the original WMD (Moverscore) and RWMD (BERTscore) as well as SBERT and WordSet-CKA on WMT 17 and 18. We report the results of RoBERTa-base and RoBERTa-large for WMT 17 and WMT 18, because they appear to be the best performing backbone models for these tasks.

To choose reasonable temperatures for TWMD and TRWMD, we tried a few values between 0.001 and 0.15 on WMT 15-16, and chose for each method based on the best averaged performance (details in Appendix \ref{sec:temperature}). The resulting temperatures for TWMD, TRWMD, TWMD-b (where ``-b" stands for batch centering of word vectors) and TRWMD-b are $T=0.02, 0.02, 0.10, 0.15$ respectively. We used a single Sinkhorn iteration for TWMD(-b). 

The main results of WMT 17 and 18 are summarized in Table \ref{table:wmt17} and \ref{table:wmt18}. Batch-mean centering appears to be helpful in improving the scores for all methods. TWMD-b performs the best in most of the cases. In particular, it is on average +2.3 / +2.8 higher than the WMD-based Moverscore-b in WMT-17 and +1.1 / +1.9 higher in WMT-18.

\begin{table}[!h]
\footnotesize
\centering
\caption{Correlation with human scores on the WMT17-18 after fine-tuning on WMT15-16. BLEURTbase and BLEURT have an extra pretraining step, as described in \cite{sellam2020bleurt}}.
\label{WMT17-ft}
\begin{tabular}{c c c}
\bf Metric & \bf \makecell[c]{\bf WMT-17\\ Avg.} & \bf \makecell[c]{\bf WMT-18\\ Avg.}\\\hline
\emph{base models} & ($\tau\ /\ r$) & ($\tau$)\\\hline
BLEURTbase -pre& 56.8 / 75.8 & 33.6\\
BLEURTbase& 61.0 / 80.2 & \textbf{34.9}\\
\makecell[c]{TWMDbase}& \bf 61.7 / 81.0 & 34.7\\
\makecell[c]{TWMDbase-b}& 61.1 / 80.7 & 34.7\\ \hline
\emph{large models} & ($\tau\ /\ r$) & ($\tau$)\\\hline
BLEURT -pre& 59.8 / 79.2 & 34.5\\
BLEURT& 62.5 / 81.8 & \textbf{35.6}\\
\makecell[c]{TWMD}& 62.8 / 81.7 & 35.5\\
\makecell[c]{TWMD-b}& \bf 63.4 / 82.9 & 35.5\\ \hline
\end{tabular}
\end{table}

\subsubsection*{Evaluation metrics with fine-tuning}
We also test the effectiveness of batch-mean centering and TWMD in the fine-tuning process. Similar to \cite{sellam2020bleurt}, we make use of the human ratings from WMT 15-16 for training, and evaluate the fine-tuned models on WMT 17 and 18. We use the L2 loss function during fine-tuning, 
\begin{align*}
     \mathrm{Loss} =& \mathrm{MSE}(\mathrm{Sim}(\Xb_1, \Xb_2), \hat{y}),
\end{align*}
where $\Xb_1$, $\Xb_2$ denotes two sentences, and $\hat{y}$ is the human score. 

We  present the result of TWMD based on the RoBERTa-base and RoBERTa-large backbones in Table \ref{WMT17-ft}. We compare the result with that of state-of-the-art BLEURT \cite{sellam2020bleurt} models.
BLEURTbase-pre and BLEURT-pre are directly fine-tuned on WMT 15-16 (with 5344 records in total), while BLEURTbase and BLEURT are additionally pretrained on a large amount of synthetic data from Wikipedia.
The scores obtained by TWMD-b not only clearly outperform BLEURTbase-pre and BLEURT-pre with the same training setting, but are comparable or better than the performance of BLEURT with the extra pretraining stage, on both base and large conditions.
This last result is especially notable considering that the synthetic data and the task setup used to further pretrain BLUERT were designed with metric similarity in mind (by leveraging on classical evaluation metrics for MT such as BLEU and ROUGE), whereas TWMD owes its performance solely to a better use of the representations.

\begin{figure}[!h]
    \centering
    \includegraphics[width=0.23\textwidth]{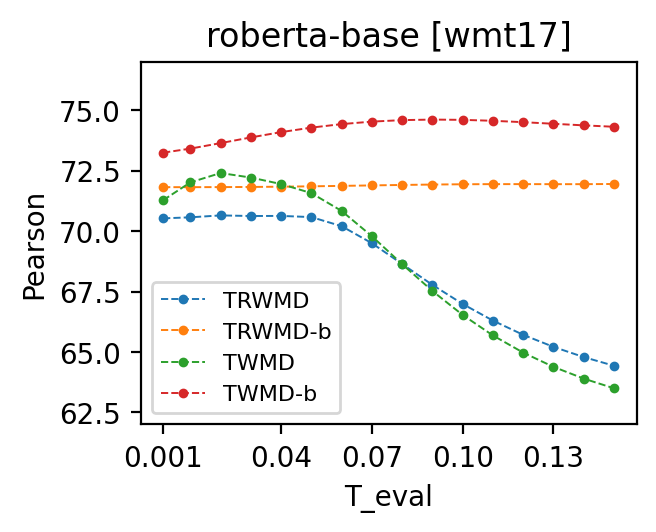}
    \includegraphics[width=0.23\textwidth]{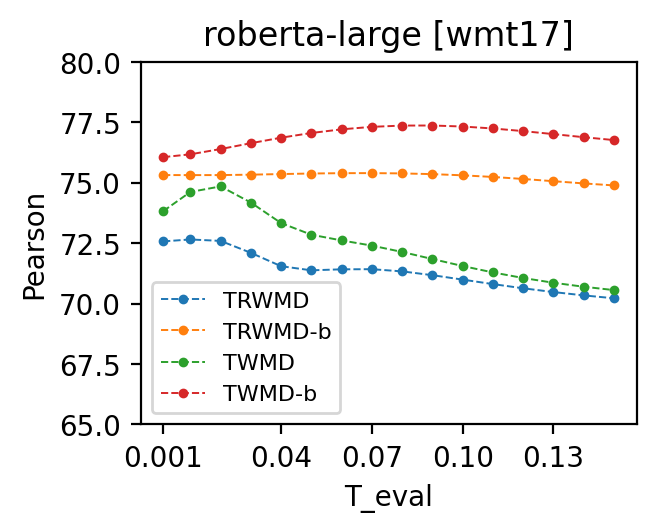}
    \includegraphics[width=0.23\textwidth]{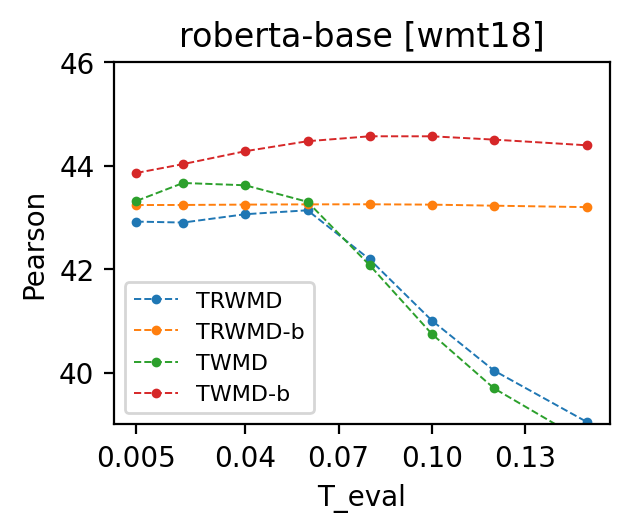}
    \includegraphics[width=0.23\textwidth]{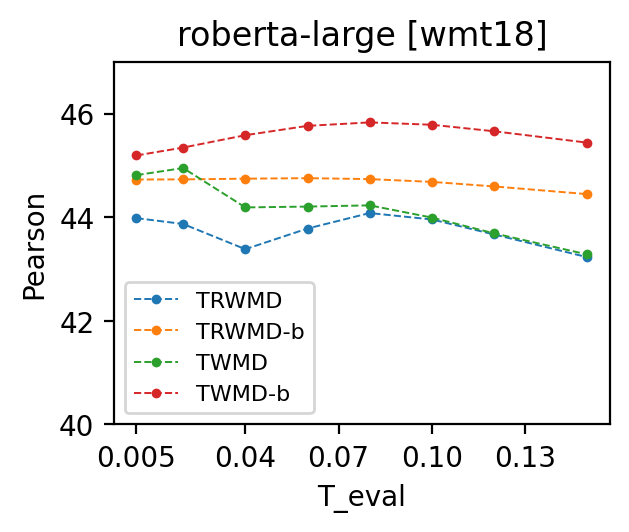}
    \caption{Pearson correlation vs. temperature in evaluation of WMT 17-18.}
    \label{fig:t_eval}
\end{figure}

\subsubsection*{Temperature dependence}
The results of TWMD, TRWMD, TWMD-b, TRWMD-b in Table \ref{table:wmt17} and \ref{table:wmt18} used the fixed temperature (tuned in WMT15-16) $T=0.02,\ 0.02,\ 0.10,\ 0.15$ for evaluation. A natural question to ask is how sensitive does the result depend on these hyperparameters. 

Figure \ref{fig:t_eval} shows the Pearson correlation vs. temperature for all four models and metrics with different temperature hyperparameters in WMT 15-18. We can see that the TWMD-b and TRWMD-b methods are robust with temperature. In comparison, TWMD and TRWMD without batch-mean centering appears sensitive to the temperature. The Kendall $\tau$ correlation follow a similar trend.

\begin{figure}[!h]
    \centering
    \includegraphics[width=0.235\textwidth]{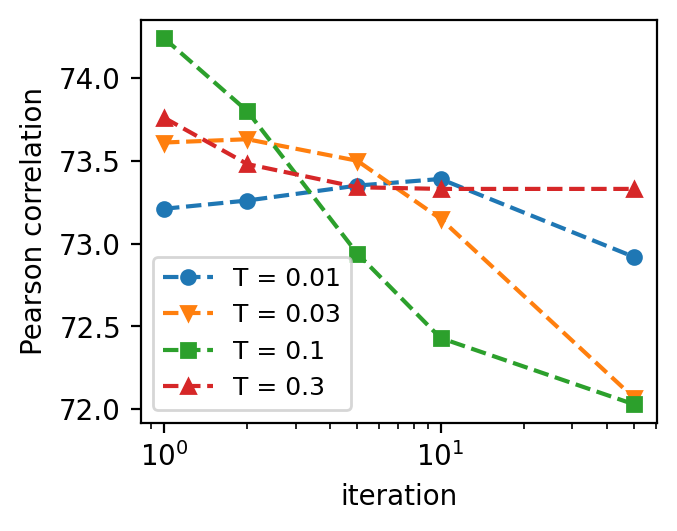}
    \includegraphics[width=0.235\textwidth]{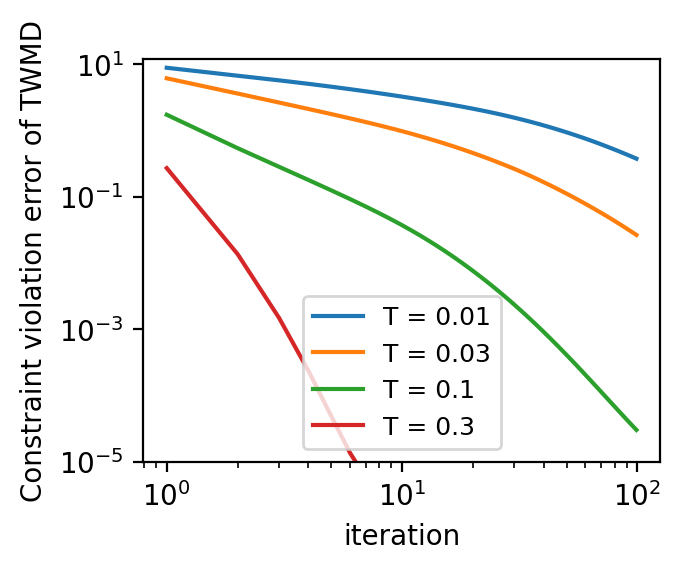}
    \caption{Left: Pearson correlation as a function of the number of iteration in Sinkhorn algorithm. Right: the convergence rate of the Sinkhorn algorithm. The underlying model in this figure is roberta-base with batch mean centered word vectors.}
    \label{fig:iteration}
\end{figure}

\subsubsection*{Sinkhorn iteration dependence for Tempered-WMD}
We also investigate how the Sinkhorn iterations affect the TWMD-b. (Figure \ref{fig:iteration}, left) shows the Pearson correlation vs the number of Sinkhorn iterations in four different temperatures. Somewhat surprisingly, although Sinkhorn algorithm needs more iterations to converge especially for low temperatures (Figure \ref{fig:iteration}, right), the Pearson correlation of TWMD with only 1 iteration is the highest\footnote{A minor exception appears to be for $T=0.01$, where the 1-iter TWMD-b is slightly worse than the 10-iter TWMD-b.} of the Sinkhorn update for various temperatures.

\section{Conclusion}
Designing automatic evaluation metrics for text is a challenging task.
Recent advances in the field leverage contextualized word representations, which are in turn generated by deep neural network models such as BERT and its variants.
We present two techniques for improving such similarity metrics:  a batch-mean centering strategy for word representations which addresses the statistical biases within deep contextualized word representations, and a computationally efficient tempered Word Mover Distance. Numerical experiments conducted using representations obtained from a range of BERT-like models confirm that our proposed metric consistently improves the correlation with human judgements.

\bibliography{main}
\bibliographystyle{acl_natbib}

\clearpage
\appendix
\section{Consistent advantage of batch centering with different layers}
\label{sec:layer}
\cite{zhang2019bertscore} did an extensive search of the best layer was done for models using WMT-16 dataset. While the best layer varies from model to model in general, the best layers for base and large versions of BERT and RoBERTa are found to be very close to the 10th and the 19th. 

\begin{figure}[!h]
    \centering
    \includegraphics[width=0.23\textwidth]{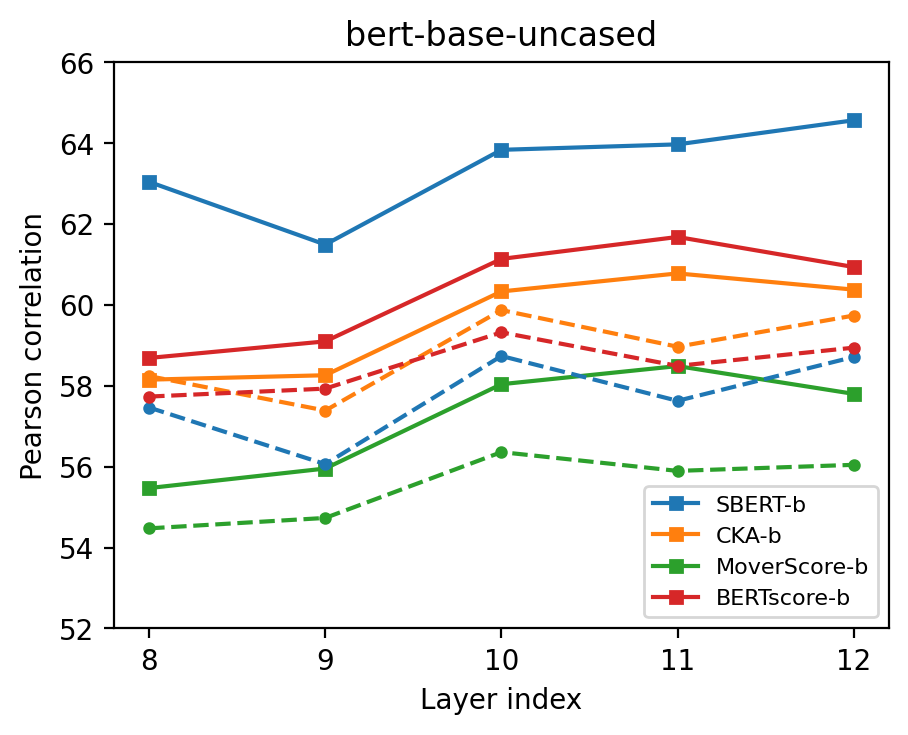}
    \includegraphics[width=0.23\textwidth]{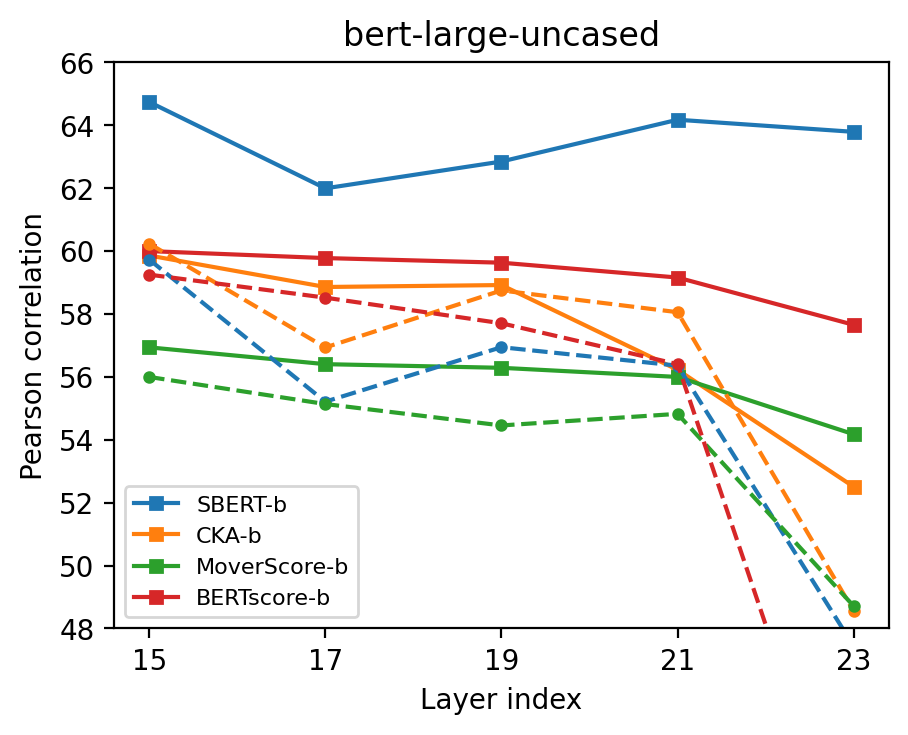}
    \includegraphics[width=0.23\textwidth]{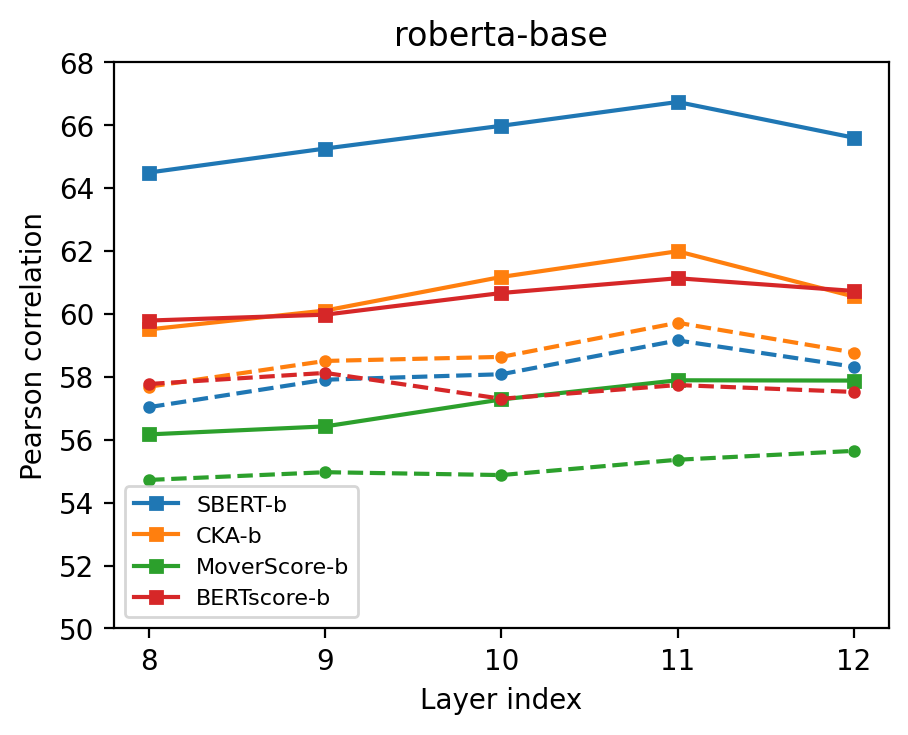}
    \includegraphics[width=0.23\textwidth]{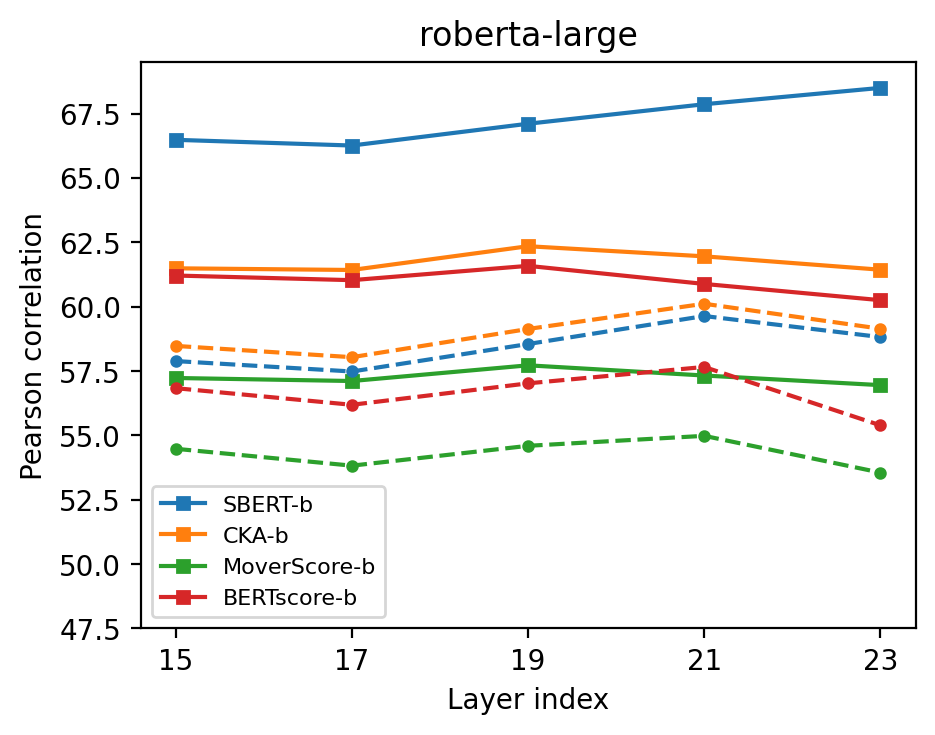}
    \caption{Four models with four metrics evaluated on STS 12-16 using different layers. Squares connected by solid lines are with batch centering. Dots connected by dashed lines are their counterparts without batch centering.}
    \label{fig:sts-layerdep}
\end{figure}

\begin{figure}[!h]
    \centering
    \includegraphics[width=0.23\textwidth]{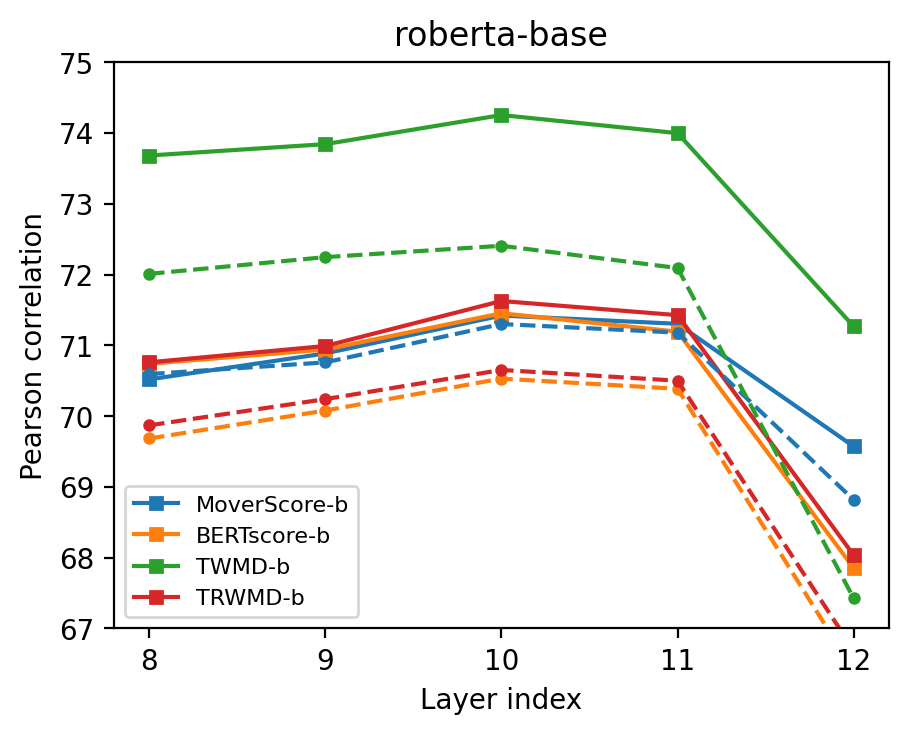}
    \includegraphics[width=0.23\textwidth]{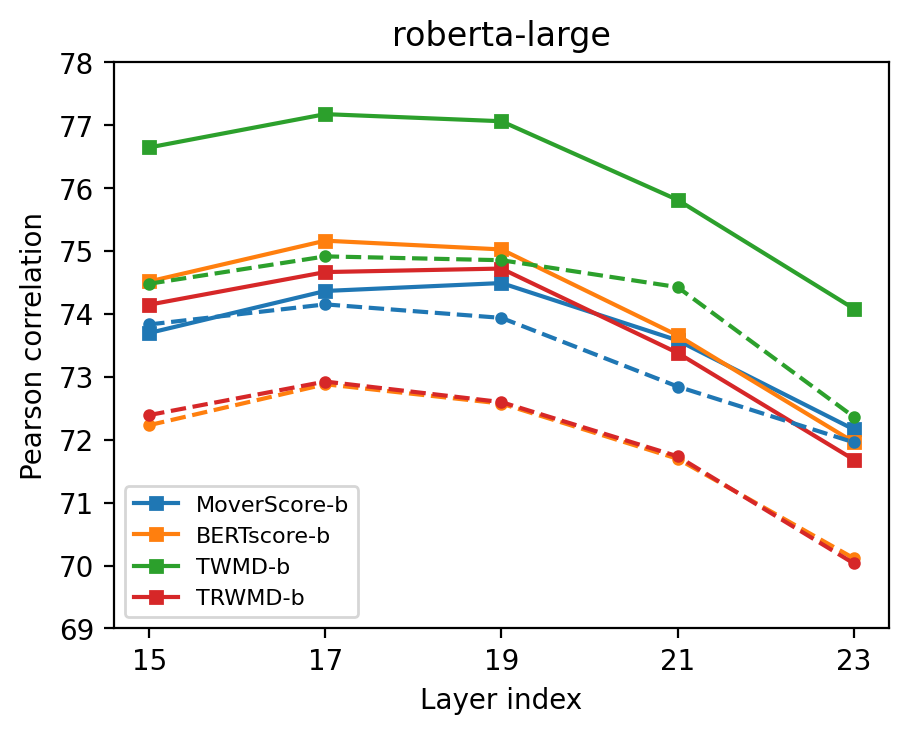}
    \caption{roberta-base and roberta-large with four metrics evaluated on WMT-17 using different layers. Squares connected by solid lines are with batch centering. Dots connected by dashed lines are their counterparts without batch centering.}
    \label{fig:wmt17-layerdep}
\end{figure}

We conduct the experiments for STS 12-16 and WMT-17 for five different layers and show the result in Fig. \ref{fig:sts-layerdep} and Fig. \ref{fig:wmt17-layerdep}. The batch centered version (solid lines) of all the metrics perform consistently better than its uncentered counterparts (dashed lines). For WMT-18, we selected two more layers for roberta-base and roberta large (Table \ref{table:wmt18-layer-dep}), where we also see the consistent advantage of batch centering. The values of Pearson correlation have little change from layer to layer as well.

\begin{table}[!ht]
\centering
\footnotesize
\caption{WMT-18 evaluated with two more layers for roberta-base and large.}
\label{table:wmt18-layer-dep}
\begin{tabular}{c c c}
\hline\hline
\bf Metric & \multicolumn{2}{c}{\makecell[c]{\bf Avg. \;\; $\tau\ /\ r$}}\\\hline\hline
\emph{roberta-base} & \bf Layer-9 & \bf Layer-11 \\\hline\hline
\makecell[c]{SBERT} & 26.2 / 36.0 & 27.1 / 37.0 \\
\makecell[c]{SBERT-b} & 29.6 / 41.7 & 29.3 / 41.3 \\\hline
\makecell[c]{CKA} & 26.3 / 36.3 & 27.1 / 37.4 \\
\makecell[c]{CKA-b} & 30.1 / 43.2 & 30.0 / 43.3 \\\hline
\makecell[c]{MoverScore} & 30.4 / 42.5 & 30.5 / 42.8 \\
\makecell[c]{MoverScore-b} & 30.5 / 42.5 & 30.5 / 42.7 \\\hline
\makecell[c]{BERTscore} & 30.0 / 42.7 & 30.2 / 42.9 \\
\makecell[c]{BERTscore-b} & 30.2 / 43.0 & 30.2 / 43.1 \\\hline
\makecell[c]{TWMD} & 30.9 / 43.6 & 30.9 / 43.5 \\
\makecell[c]{TWMD-b} & 31.5 / 44.3 & 31.4 / 44.4 \\\hline
\makecell[c]{TRWMD} & 30.1 / 42.7 & 30.3 / 42.8 \\
\makecell[c]{TRWMD-b} & 30.5 / 42.9 & 30.6 / 43.1 \\
\hline\hline

\emph{roberta-large} & \bf Layer-17 & \bf Layer-21\\\hline\hline
\makecell[c]{SBERT} & 29.6 / 40.5 & 28.7 / 40.4 \\
\makecell[c]{SBERT-b} & 31.4 / 43.7 & 30.4 / 42.5 \\\hline
\makecell[c]{CKA} & 29.6 / 40.9 & 28.8 / 40.7 \\
\makecell[c]{CKA-b} & 31.4 / 45.0 & 30.8 / 44.3 \\\hline
\makecell[c]{MoverScore} & 31.8 / 43.9 & 31.3 / 43.1 \\
\makecell[c]{MoverScore-b} & 31.6 / 43.7 & 31.3 / 43.3 \\\hline
\makecell[c]{BERTscore} & 31.8 / 44.1 & 30.8 / 43.2 \\
\makecell[c]{BERTscore-b} & 31.6 / 44.8 & 31.0 / 44.0 \\\hline
\makecell[c]{TWMD} & 32.5 / 44.9 & 31.8 / 44.5 \\
\makecell[c]{TWMD-b} & 32.7 / 45.9 & 32.1 / 45.1 \\\hline
\makecell[c]{TRWMD} & 31.7 / 44.0 & 30.8 / 43.2 \\
\makecell[c]{TRWMD-b} & 31.8 / 44.4 & 31.2 / 43.7 \\
\hline\hline

\end{tabular}
\end{table}

\begin{figure}[!h]
    \centering
    \includegraphics[width=0.23\textwidth]{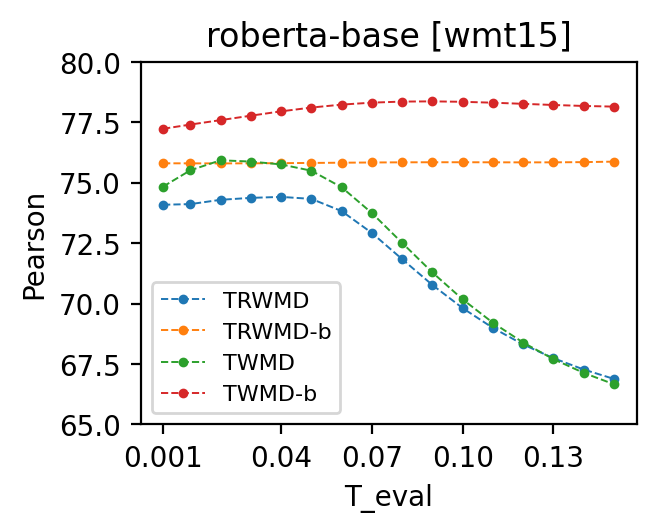}
    \includegraphics[width=0.23\textwidth]{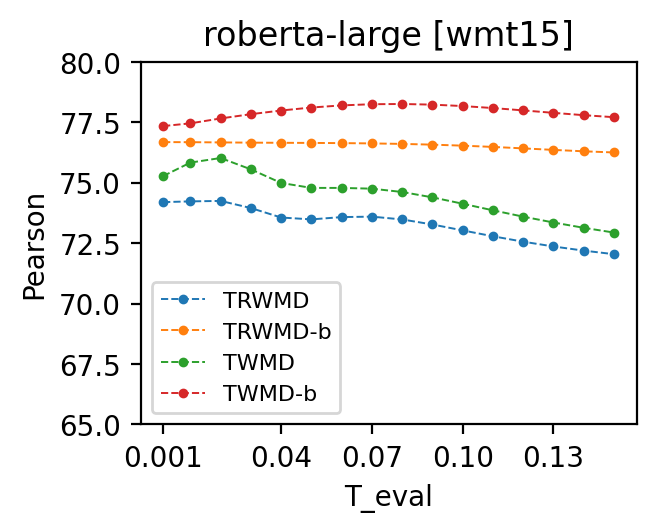}
    \includegraphics[width=0.23\textwidth]{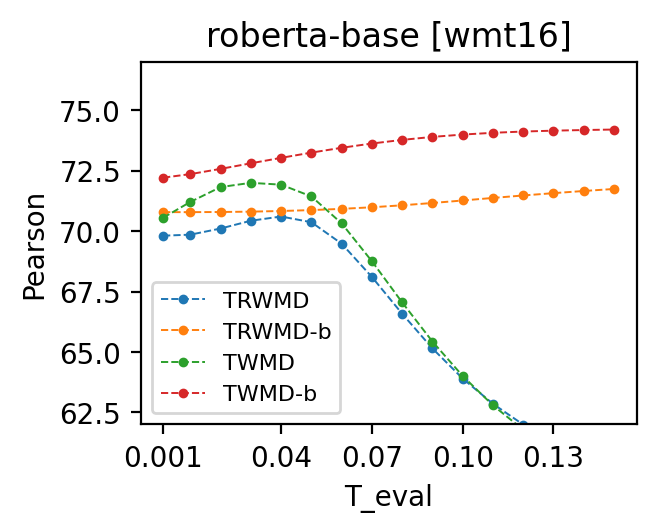}
    \includegraphics[width=0.23\textwidth]{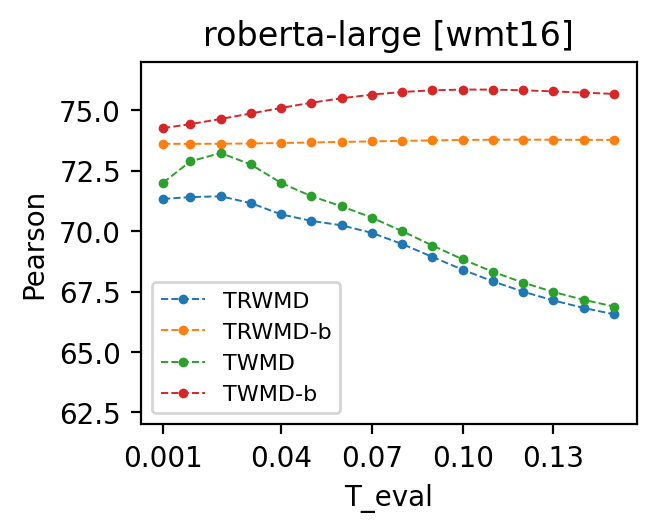}
    \caption{Pearson correlation vs. temperature in evaluation of WMT 15-16.}
\end{figure}

\section{Choice of temperature using WMT-15 and 16 as validation}
\label{sec:temperature}

We use the datasets WMT-15 and WMT-16 to determine the optimal temperature for TWMD, TRWMD, TWMD-b and TRWMD-b, then fix the temperature for evaluating WMT-17 (Table \ref{table:wmt17}) and WMT-18 (Table \ref{table:wmt18}).

In particular, we first estimate the Pearson correlation for each metric with roberta-base and roberta-large on WMT-15 and 16. Next we compute the weighted mean of WMT-15 and 16 by $\bar{r} = (N_{15}*r_{15}+N_{16}*r_{16})/(N_{15}+N_{16})$ where $N$ and $r$ is the size of dataset and the Pearson correlation. Last, we take the mean of $\bar{r}$ for roberta-base and roberta-large and use the result $r = (\bar{r}_\mathrm{base} + \bar{r}_\mathrm{large})/2$ to determine the optimal temperature for each metric. We thus obtained the best temperature for TWMD and TRWMD to be 0.02, and the best temperatures for TWMD-b and TRWMD-b to be 0.1 and 0.15, respectively.

\section{Precision and F1 Scores}
\label{sec:f1}
The BERTscore-Recall (Eq. \ref{eq:bertscore}) is an asymmetic metric. \cite{zhang2019bertscore} also provides two additional metrics: the BERTscore-Precision that switches the roles of the query and reference sentences; and a symmetric BERTscore-F1 metric that is the harmonic mean of BERTscore-Recall and BERTscore-Precision metrics. Since BERTscore-F1 is an ensemble of the two metrics, it is expected to perform better. However, according to~\cite{zhang2019bertscore}, BERTscore-Precision and BERTscore-F1 are inconsistent and sometimes significantly underperform (e.g. in COCO image captioning). 

Similar to BERTscore, both TRWMD and TWMD metrics are asymmetric\footnote{TWMD is asymmetric when only few Sinkhorn iterations are applied.}. In Table~\ref{table:wmt17-precision} and \ref{table:wmt18-precision}, we present the results of the precision-scores of BERTscore~\cite{zhang2019bertscore}, TRWMD and TWMD with or without batch centering in WMT-17 and 18, where the temperatures are the same as those for the recall-scores (see Appendix \ref{sec:temperature}). In Table~\ref{table:wmt17-f1} and \ref{table:wmt18-f1}, we present the results of the F1-scores, where the temperature for TWMD, TRWMD, TWMD-b and TRWMD-b are 0.01, 0.01, 0.08 and 0.06 respectively. TWMD-b is the top performer in most of cases, albeit the lead shrinks in the cases of the F1-scores. In addition, batch-centering produces consistent improvements for all metrics.

\begin{table*}[!h]
\scriptsize
\centering
\caption{Correlation with human scores on the WMT-17 Metrics Shared Task using the Precision scores for BERTscore, TWMD and TRWMD with or without batch centering. }
\label{table:wmt17-precision}
\begin{tabular}{c c c c c c c c c}
\hline
\bf Metric & \bf \makecell[c]{\bf cs-en\\$\tau\ /\ r$} & \bf \makecell[c]{\bf de-en\\$\tau\ /\ r$} & \bf \makecell[c]{\bf fi-en\\$\tau\ /\ r$} & \bf \makecell[c]{\bf lv-en\\$\tau\ /\ r$} & \bf \makecell[c]{\bf ru-en\\$\tau\ /\ r$} & \bf \bf \makecell[c]{\bf tr-en\\$\tau\ /\ r$} & \bf \makecell[c]{\bf zh-en\\$\tau\ /\ r$} & \bf \makecell[c]{\bf Avg.\\$\tau\ /\ r$}\\\hline
\emph{roberta-base}\\\hline
\makecell[c]{BERTscore} & 47.9 / 66.1 & 47.5 / 66.0 & 56.9 / 76.9 & 48.0 / 67.1 & 50.7 / 68.6 & 51.3 / 70.3 & 53.5 / 74.6 & 50.8 / 70.0\\
\makecell[c]{BERTscore-b} & 48.6 / 67.8 & 48.5 / 67.7 & 58.5 / 78.7 & 48.1 / 68.5 & 51.1 / 70.3 & 53.8 / 73.1 & 52.3 / 73.8 & 51.6 / 71.3\\
\makecell[c]{TWMD} & 48.8 / 66.4 & 49.2 / 67.9 & 61.9 / 80.9 & 51.0 / 70.3 & 52.6 / 71.8 & 54.0 / 73.6 & \bf 55.0 / 75.9 & 53.2 / 72.3\\
\makecell[c]{TWMD-b} & \bf 50.8 / 69.5 & \bf 51.7 / 70.8 & \bf 63.4 / 83.1 & \bf 52.4 / 72.5 & \bf 54.0 / 73.9 & \bf 56.8 / 77.1 & 54.3 / 75.4 & \bf 54.8 / 74.6\\
\makecell[c]{TRWMD} & 47.9 / 65.9 & 47.6 / 66.5 & 57.0 / 77.0 & 48.0 / 67.0 & 50.8 / 68.8 & 51.4 / 71.1 & 53.4 / 74.5 & 50.9 / 70.1\\
\makecell[c]{TRWMD-b} & 49.1 / 67.7 & 49.7 / 68.5 & 58.6 / 79.3 & 48.5 / 68.2 & 52.0 / 70.7 & 54.3 / 74.6 & 52.4 / 73.5 & 52.1 / 71.8\\\hline 

\emph{roberta-large}\\\hline
\makecell[c]{BERTscore} & 51.9 / 69.5 & 54.3 / 72.8 & 57.3 / 77.2 & 51.3 / 70.1 & 54.6 / 71.7 & 53.1 / 72.5 &
 56.2 / 76.1 & 54.1 / 72.8\\
\makecell[c]{BERTscore-b} & 52.2 / 72.0 & 54.3 / 73.5 & 60.0 / 80.0 & 52.2 / 72.8 & 55.4 / 74.4 & 56.2 / 75.4
 & 56.0 / 76.7 & 55.2 / 75.0\\
\makecell[c]{TWMD} & 52.2 / 69.0 & 55.5 / 74.0 & 62.7 / 81.2 & 54.0 / 72.8 & 56.0 / 74.4 & 56.1 / 74.7 & \bf 57.9
 / 78.0 & 56.3 / 74.9\\
\makecell[c]{TWMD-b} & \bf 54.6 / 73.9 & \bf 56.8 / 76.0 & \bf 64.5 / 83.6 & \bf 55.7 / 75.5 & \bf 57.4 / 76.5 & \bf 58.1 / 78.3 & 57.4 / \bf 78.0 & \bf 57.8 / 77.4\\
\makecell[c]{TRWMD} & 51.5 / 68.8 & 54.3 / 72.7 & 56.8 / 76.9 & 51.0 / 69.8 & 54.5 / 71.7 & 52.8 / 72.5 & 56.2 / 76.0 & 53.9 / 72.6\\
\makecell[c]{TRWMD-b} & 52.4 / 71.3 & 54.7 / 73.5 & 59.4 / 79.7 & 52.0 / 71.8 & 55.3 / 73.4 & 55.7 / 76.0 & 55.6 / 76.0 & 55.0 / 74.5\\
\hline
\end{tabular}
\end{table*}

\begin{table*}[!h]
\scriptsize
\centering
\caption{Correlation with human scores on the WMT-18 Metrics Shared Task using the Precision scores for BERTscore, TWMD and TRWMD with or without batch centering. }
\label{table:wmt18-precision}
\begin{tabular}{c c c c c c c c c}
\hline
\bf Metric & \bf \makecell[c]{\bf cs-en\\$\tau\ /\ r$} & \bf \makecell[c]{\bf zh-en\\$\tau\ /\ r$} & \bf \makecell[c]{\bf ru-en\\$\tau\ /\ r$} & \bf \makecell[c]{\bf fi-en\\$\tau\ /\ r$} & \bf \makecell[c]{\bf tr-en\\$\tau\ /\ r$} & \bf \bf \makecell[c]{\bf et-en\\$\tau\ /\ r$} & \bf \makecell[c]{\bf de-en\\$\tau\ /\ r$} & \bf \makecell[c]{\bf Avg.\\$\tau\ /\ r$}\\\hline
\emph{roberta-base}\\\hline

\makecell[c]{BERTscore} & 28.9 / 40.4 & 27.7 / 37.8 & 27.3 / 39.4 & 25.2 / 36.5 & 30.8 / 43.1 & 33.9 / 48.1 &
 39.1 / 55.0 & 30.4 / 42.9\\
\makecell[c]{BERTscore-b} & 29.2 / 41.6 & 27.5 / 38.1 & 27.5 / 39.9 & 25.4 / 37.2 & 30.9 / 43.2 & 34.2 / 48.9
 & 39.4 / 55.6 & 30.5 / 43.5\\
\makecell[c]{TWMD} & 29.2 / 41.3 & 28.4 / 38.4 & 28.1 / 40.2 & 25.6 / 36.7 & 31.5 / 43.6 & 34.8 / 49.3 & 39.9
 / 56.4 & 31.1 / 43.7\\
\makecell[c]{TWMD-b} & \bf 29.7 / 42.3 & \bf 28.7 / 39.1 & \bf 28.7 / 40.9 & \bf 26.4 / 38.3 & \bf 32.2 / 44.4 & \bf 35.4 / 50.3 & \bf 40.4 / 57.3 & \bf 31.6 / 44.7\\
\makecell[c]{TRWMD} & 28.9 / 40.4 & 27.7 / 37.8 & 27.4 / 39.4 & 25.1 / 36.1 & 30.9 / 43.0 & 33.9 / 48.1 & 39.2 / 55.1 & 30.4 / 42.9\\
\makecell[c]{TRWMD-b} & 29.4 / 41.6 & 27.9 / 37.9 & 28.1 / 39.9 & 25.3 / 36.6 & 31.2 / 43.0 & 34.5 / 49.1 & 39.8 / 55.9 & 30.9 / 43.4\\
\hline

\emph{roberta-large}\\\hline
\makecell[c]{BERTscore} & 30.0 / 41.6 & 28.2 / 38.3 & 29.0 / 40.9 & 26.7 / 37.8 & 31.6 / 43.6 & 35.9 / 49.8 & 41.0 / 57.4 & 31.8 / 44.2\\
\makecell[c]{BERTscore-b} & 30.5 / 43.7 & 28.1 / 38.7 & 28.9 / 41.3 & 27.1 / 39.1 & 31.5 / 44.2 & 35.9 / 50.6 & 41.0 / 57.5 & 31.9 / 45.0\\
\makecell[c]{TWMD} & 30.7 / 42.9 & 28.9 / 39.1 & 29.5 / 41.1 & 27.3 / 38.5 & 32.3 / 44.3 & 36.6 / 50.7 & 41.8 / 58.6 & 32.5 / 45.1\\
\makecell[c]{TWMD-b} & \bf 31.3 / 44.4 & \bf 29.0 / 39.5 & \bf 29.9 / 41.9 & \bf 27.7 / 39.8 & \bf 32.6 / 45.0 & \bf 36.8 / 51.5 & \bf 41.9 / 59.0 & \bf 32.8 / 45.9\\
\makecell[c]{TRWMD} & 29.7 / 41.3 & 28.2 / 38.2 & 29.0 / 40.6 & 26.5 / 37.4 & 31.6 / 43.4 & 35.8 / 49.7 & 41.0 / 57.3 & 31.7 / 44.0\\
\makecell[c]{TRWMD-b} & 30.4 / 43.1 & 28.2 / 38.3 & 29.2 / 41.0 & 26.7 / 38.2 & 31.7 / 43.7 & 35.9 / 50.4 & 41.1 / 57.5 & 31.9 / 44.6\\
\hline

\end{tabular}
\end{table*}

\begin{table*}[!h]
\scriptsize
\centering
\caption{Correlation with human scores on the WMT-17 Metrics Shared Task using the F1 scores for BERTscore, TWMD and TRWMD with or without batch centering. }
\label{table:wmt17-f1}
\begin{tabular}{c c c c c c c c c}
\hline
\bf Metric & \bf \makecell[c]{\bf cs-en\\$\tau\ /\ r$} & \bf \makecell[c]{\bf de-en\\$\tau\ /\ r$} & \bf \makecell[c]{\bf fi-en\\$\tau\ /\ r$} & \bf \makecell[c]{\bf lv-en\\$\tau\ /\ r$} & \bf \makecell[c]{\bf ru-en\\$\tau\ /\ r$} & \bf \bf \makecell[c]{\bf tr-en\\$\tau\ /\ r$} & \bf \makecell[c]{\bf zh-en\\$\tau\ /\ r$} & \bf \makecell[c]{\bf Avg.\\$\tau\ /\ r$}\\\hline
\emph{roberta-base}\\\hline
\makecell[c]{BERTscore} & \textbf{50.2} / 68.8 & 50.3 / 69.3 & 62.9 / 82.0 & 51.3 / 71.1 & 53.0 / 72.1 & 54.6 / 74.0 & 54.4 / 75.5 & 53.8 / 73.3\\
\makecell[c]{BERTscore-b} & \bf 50.2 / 69.3 & 51.0 / 70.5 & 62.9 / 82.3 & 50.9 / 71.8 & 53.0 / 72.8 & 55.4 / 75.2 & 52.9 / 74.4 & 53.8 / 73.8\\

\makecell[c]{TWMD} & 48.4 / 66.3 & 49.5 / 68.5 & 62.3 / 81.0 & 51.5 / 70.9 & 52.6 / 72.0 & 54.5 / 73.7 & \textbf{55.2 / 76.0} & 53.4 / 72.6\\
\makecell[c]{TWMD-b} & 50.0 / 68.8 & \bf 51.4 / 70.8 & \bf 63.1 / 82.9 & \bf 52.1 / 72.5 & \textbf{53.6 / 73.6} & \textbf{56.5 / 76.6} & 54.1 / 75.4 &\bf 54.4 / 74.4\\

\makecell[c]{TRWMD} & \textbf{50.2} / 68.8 & 50.4 / 69.3 & 63.0 / 82.0 & 51.4 / 71.1 & 53.1 / 72.1 & 54.7 / 74.1 & 54.5 / 75.6 & 53.9 / 73.3\\
\makecell[c]{TRWMD-b} & \bf 50.2 / 69.3 & 51.1 / 70.6 & 63.0 / 82.3 & 51.0 / 71.8 & 53.2 / 73.1 & 55.6 / 75.4 & 53.3 / 74.6 & 53.9 / 73.9\\
\hline

\emph{roberta-large}\\\hline
\makecell[c]{BERTscore} & 54.0 / 72.0 & 56.2 / 75.0 & 63.1 / 81.8 & 54.8 / 73.5 & 56.2 / 73.9 & 56.3 / 75.4 & 57.4 / 77.6 & 56.8 / 75.6\\
\makecell[c]{BERTscore-b} & 54.1 / \bf 74.0 & 56.2 / 75.9 & {\bf 64.6} / 83.5 & 55.3 / 75.9 & 56.8 / 76.2 & 57.4 / 77.0 & 56.4 / 77.3 & 57.3 / 77.1\\

\makecell[c]{TWMD} & 52.5 / 69.3 & 55.8 / 74.5 & 63.5 / 81.5 & 54.6 / 73.5 & 56.2 / 74.6 & 56.6 / 74.8 & \bf 58.0 / 78.2 & 56.7 / 75.2\\

\makecell[c]{TWMD-b} & 54.1 / 73.5 & 56.2 / 75.8 & \bf 64.6 / 83.6 & 55.4 / 75.6 & \bf 57.2 / 76.7 & \bf 58.0 / 77.8 & 57.1 / 77.8 & \bf 57.5 / 77.3\\

\makecell[c]{TRWMD} & 54.0 / 72.1 & \textbf{56.3} / 75.1 & 63.1 / 81.8 & 54.8 / 73.5 & 56.3 / 74.0 & 56.3 / 75.4 & 57.5 / 77.7 & 56.8 / 75.7\\
\makecell[c]{TRWMD-b} & \bf 54.3 / 74.0 & \bf 56.3 / 76.0 & {\bf 64.6} / 83.5 & \bf 55.5 / 76.0 & 56.9 / 76.4 & 57.5 / 77.2 & 56.8 / 77.5 & 57.4 / 77.2\\
\hline
\end{tabular}
\end{table*}

\begin{table*}[!h]
\scriptsize
\centering
\caption{Correlation with human scores on the WMT-18 Metrics Shared Task using the F1 scores for BERTscore, TWMD and TRWMD with or without batch centering. }
\label{table:wmt18-f1}
\begin{tabular}{c c c c c c c c c}
\hline
\bf Metric & \bf \makecell[c]{\bf cs-en\\$\tau\ /\ r$} & \bf \makecell[c]{\bf zh-en\\$\tau\ /\ r$} & \bf \makecell[c]{\bf ru-en\\$\tau\ /\ r$} & \bf \makecell[c]{\bf fi-en\\$\tau\ /\ r$} & \bf \makecell[c]{\bf tr-en\\$\tau\ /\ r$} & \bf \bf \makecell[c]{\bf et-en\\$\tau\ /\ r$} & \bf \makecell[c]{\bf de-en\\$\tau\ /\ r$} & \bf \makecell[c]{\bf Avg.\\$\tau\ /\ r$}\\\hline
\emph{roberta-base}\\\hline
\makecell[c]{BERTscore} & \textbf{29.5} / 41.9 & 28.4 / 39.1 & 28.4 / \bf 41.0 & 26.1 / 37.9 & 31.9 / \bf 44.6 & 35.0 / 49.5 & 40.2 / 56.9 & 31.4 / 44.4\\
\makecell[c]{BERTscore-b} & \bf 29.5 / 42.1 & 28.1 / 39.1 & 28.4 / \bf 41.0 & \bf 26.4 / 38.6 & 31.9 / \bf 44.6 & 35.1 / 49.9 & 40.3 / 57.0 & 31.4 / 44.6\\
\makecell[c]{TWMD} & 29.0 / 41.1 & 28.4 / 38.5 & 28.1 / 40.2 & 25.6 / 36.8 & 31.7 / 43.9 & 34.9 / 49.4 & 40.0 / 56.6 & 31.1 / 43.8\\
\makecell[c]{TWMD-b} & \bf 29.5 / 42.1 & \textbf{28.5} / 39.0 & \textbf{28.7} / 40.8 & \textbf{26.4} / 38.3 & \textbf{32.2} / 44.5 & \bf 35.5 / 50.3 & \bf 40.4 / 57.4 & \bf 31.7 / 44.7\\
\makecell[c]{TRWMD} & \textbf{29.5} / 41.8 & 28.4 / 39.1 & 28.5 / \bf 41.0 & 26.1 / 37.8 & 31.9 / \bf 44.6 & 35.0 / 49.5 & 40.2 / 56.9 & 31.4 / 44.4\\
\makecell[c]{TRWMD-b} & \textbf{29.5 / 42.1} & 28.4 / \bf 39.2 & 28.4 / 40.9 & \bf 26.4 / 38.6 & 32.0 / \textbf{44.6} & 35.1 / 49.9 & 40.3 / 57.1 & 31.5 / 44.6\\

\hline

\emph{roberta-large}\\\hline
\makecell[c]{BERTscore} & 30.8 / 43.3 & 28.9 / 39.3 & \textbf{29.9} / 41.6 & 27.4 / 38.7 & 32.4 / 44.7 & 36.7 / 50.6 & \textbf{42.0 / 59.0} & 32.6 / 45.3\\
\makecell[c]{BERTscore-b} & 31.2 / \textbf{44.6} & 28.9 / \textbf{39.8} & 29.7 / \textbf{42.4} & \textbf{27.8 / 40.3} & \textbf{32.5} / 45.4 & \textbf{36.8} / 51.2 & 41.8 / \textbf{59.0} & \bf 32.7 / 46.1\\
\makecell[c]{TWMD-b} & 30.8 / 43.2 & \textbf{29.0} / 39.4 & 29.5 / 41.3 & 27.5 / 38.8 & \textbf{32.5} / 44.7 & 36.7 / 50.7 & 41.8 / 58.9 & 32.6 / 45.3\\
\makecell[c]{TWMD-b} & \textbf{31.3} / 44.5 & \textbf{29.0} / 39.7 & 29.7 / 42.0 & 27.7 / 39.9 & \textbf{32.5} / 45.2 & \bf 36.8 / 51.4 & 41.8 / \textbf{59.0} & \textbf{32.7} / 46.0\\
\makecell[c]{TRWMD} & 30.8 / 43.3 & \textbf{29.0} / 39.3 & \textbf{29.9} / 41.6 & 27.4 / 38.7 & 32.4 / 44.7 & 36.7 / 50.6 & \textbf{42.0 / 59.0} & 32.6 / 45.3\\
\makecell[c]{TRWMD-b} & 31.2 / \textbf{44.6} & 28.9 / \textbf{39.8} & 2\textbf{9.9 / 42.4} & \textbf{27.8 / 40.3} & \textbf{32.5 / 45.5} & \textbf{36.8 }/ 51.1 & 41.8 / \textbf{59.0} & \bf 32.7 / 46.1\\
\hline

\end{tabular}
\end{table*}

\end{document}